\documentclass[1p, authoryear]{elsarticle}

\usepackage[colorlinks=True]{hyperref}
\usepackage{url}
\usepackage{gensymb}
\usepackage{amsthm}
\usepackage{amsmath}
\usepackage{thmtools}
\usepackage{thm-restate}
\usepackage{bbm}
\usepackage{dutchcal}
\usepackage{graphicx}
\usepackage{cleveref}
\usepackage{amssymb}
\usepackage{bbm}
\usepackage{mathtools}
\usepackage{xcolor}
\usepackage{ulem}
\usepackage{xpatch}
\usepackage{tikz-cd}
\usepackage{float}
\usepackage{tabularx}
\usepackage[font=small, margin=20pt]{caption}

\newcommand{\I}{\mathbbm{1}}

\newcommand{\N}{\mathbb{N}}

\newcommand{\Z}{\mathbb{Z}}

\makeatletter
\gdef\emailauthor#1#2{%
  \stepcounter{ead}%
  \g@addto@macro\@elseads{%
    \raggedright
    \let\corref\@gobble
    \def\@@tmp{#1}%
    \eadsep{\ttfamily\expandafter\strip@prefix\meaning\@@tmp}%
    \def\eadsep{\unskip,\space}%
  }%
}
\makeatother

\begin{document}

\begin{frontmatter}
\title{A Group Theoretic Analysis of the Symmetries Underlying Base Addition and Their Learnability by Neural Networks}

\author[1]{Cutter Dawes\corref{cor1}}
    \ead{cdawes@princeton.edu}
\author[2]{Simon Segert}
\author[3]{Kamesh Krishnamurthy}
\author[4,5]{Jonathan D. Cohen}
\cortext[cor1]{Corresponding author}
\affiliation[1]{organization={Department of Mathematics,
    Princeton University},
    orp={}
}
\affiliation[2]{
    organization={Independent researcher},
    orp={}
}
\affiliation[3]{organization={Zyphra},
    orp={}
}
\affiliation[4]{organization={Princeton Neuroscience Institute,
    Princeton University},
    orp={}
}
\affiliation[5]{organization={Department of Psychology,
    Princeton University},
    orp={}
}

\begin{abstract}
    A major challenge in the use of neural networks both for modeling human cognitive function and for artificial intelligence is the design of systems with the capacity to efficiently learn functions that support radical generalization.
    At the roots of this is the capacity to discover and implement symmetry functions.
    In this paper, we investigate a paradigmatic example of radical generalization through the use of symmetry: base addition.
    We present a group theoretic analysis of base addition, a fundamental and defining characteristic of which is the carry function -- the transfer of the remainder, when a sum exceeds the base modulus, to the next significant place.
    Our analysis exposes a range of alternative carry functions for a given base, and we introduce quantitative measures to characterize these. 
    We then exploit differences in carry functions to probe the inductive biases of neural networks in symmetry learning, by training neural networks to carry out base addition using different carries, and comparing efficacy and rate of learning as a function of their structure.
    We find that even simple neural networks can achieve radical generalization with the right input format and carry function, and that learnability is closely correlated with carry function structure.
    We then discuss the relevance this has for cognitive science and machine learning.
\end{abstract}

\end{frontmatter}

\section{Introduction} \label{sec:1}

One of the defining characteristics of human cognitive function, that continues to distinguish it from the most advanced systems in machine learning and artificial intelligence, is the efficiency with which people can acquire and represent knowledge that generalizes widely out of distribution.
This can be distilled, in the purest form, to the discovery of symmetries: forms of structure that can be identified in one domain, and that remain invariant over a relevant set of transformations into others.
It has been said that ``it is only slightly overstating the case to say that physics'' -- one of the most fundamental and widely applicable domains of human knowledge -- ``is the study of symmetry'' \citep{Anderson1978}.

The same might be said about the understanding of human cognition, as well as efforts to approximate its capacity for sample efficient learning and radical generalization in artificial systems.
This is because, where the structure of a data distribution is defined by a symmetry, 
then the distribution can be learned with only the amount of data needed to infer the symmetry, which can then be used for effective generalization not only over any additional data within the envelope of the training distribution, but all data that exhibits the same structure outside of the distribution as well.
This follows from the definition of a symmetry: that it is invariant to transformations of the data that conform to its structure.

Neural networks have been developed that, with the appropriate inductive bias, can efficiently discover symmetries in specific domains.
In so doing, these models can match or exceed human performance in those domains.
For example, convolutional neural networks are imbued with mechanisms for ``weight sharing'' that allow them to discover spatially invariant constituents of visual displays, such as human recognizable objects, and then identify such objects even when they appear in previously unseen locations \citep{LeCun1998, Krizhevsky2012}.
Here, the nature and scope of the symmetry is well defined -- translational invariance over spatial position -- and explicitly implemented in the model (as the spatial convolution operation).
Similarly, since recurrent networks use the same weights to process all elements of a sequence, they can be viewed as implementing of a form of ``weight sharing over time,'' which may explain their ability to discover symmetries in sequential structure in ways similar to people \citep{cleeremans1991learning, segert2022a, Elman1990, Ji-An2023}.
More powerful architectures, such as transformers, also appear to be capable of discovering symmetries that can be remarkably abstract.
This is evidenced by the impressive generalization capabilities of large language models (LLMs), such as their ability to solve analogical reasoning problems at or above human level performance \citep[e.g.,][]{Webb2023}.
In some cases, the nature of the symmetries are coming into view that appear to support simple forms of symbolic computation \citep[e.g.,][]{Yang2025}.
However, in many other cases it is not clear precisely what symmetries have been discovered, and in virtually all cases the amount of data needed for the system to discover them is considerably greater than humans require to exhibit similar generalization capabilities.

As a step towards understanding the capabilities described above, we focus on one of the simplest and most fundamental forms of symmetry: base arithmetic.
The ability to count algorithmically using a base (most commonly 10) is one of the earliest milestones of human learning \citep{Wynn1992, Dehaene1997, Piantadosi2023}, and is fundamental to base arithmetic, which lies at the heart of mathematical reasoning, our most abstract form of thought and communication.
Despite this fact, little work has been done characterizing base arithmetic from a group theoretic perspective.
Furthermore, to date, there are no neural networks that have been shown to achieve, strictly through learning, demonstrably algorithmic forms of base arithmetic computation.
Here, we seek to make progress toward both of these closely related goals: a deeper understanding of the symmetry properties of base arithmetic; and an examination of how this relates to the ability of neural networks to learn such symmetries.

Specifically, we focus on the symmetries inherent in base addition.
This is of course an elemental component of base arithmetic, and provides a simple and well defined example of symmetry: once the operations needed to perform addition are known for some subset of the data (e.g., for a number of digits, or ``places'', sufficient to reveal the underlying carry function), they can be applied to indefinitely large numbers. 
In the sections that follow, we begin by providing a formal group theoretic analysis of the operations underlying base addition, and then examine the ability of neural networks to learn these operations.
The group theoretic analysis allows us to identify, for a given base, different symmetry functions (each corresponding to a different carry function) that are functionally equivalent, but vary in structure.
We  describe several quantitative measures (fractal dimension, frequency of carrying, and associativity fraction) that we use to characterize the structure of each set of carry functions, and find that these quantitative measures are tightly correlated with how efficiently and effectively neural networks learn those carry functions.
We suggest that these observations may provide a normative account of the form of base addition people use, that involves the simplest symmetries; and that this, in turn, may be useful in designing neural networks capable of learning and exploiting such symmetry functions in the service of fully algorithmic and generalizable base addition.

Section~\ref{sec:2} provides a formal group theoretic analysis of base addition (including an overview of necessary group theoretic concepts in Section~\ref{sec:2.1}, which can be skipped by readers familiar with these concepts).
Section~\ref{sec:3} introduces quantitative measures that characterize carry function structure, and Section~\ref{sec:4} reports results from training neural networks to add using each carry function.
Section~\ref{sec:5} provides a general discussion of the relevance of our findings to related questions in mathematics, neural networks, and cognitive science, and Section~\ref{sec:6} provides a conclusion.
\section{Mathematics of Symmetry and Base Addition} \label{sec:2}

\subsection{Symmetry in Group Theory} \label{sec:2.1}

The branch of algebra known as group theory provides a formal definition of symmetry in terms of groups.
A \textit{group} is a set $G$ together with a group operation $\cdot$ satisfying the group axioms:
\begin{enumerate}
    \setlength{\itemsep}{0pt}
    \setlength{\parskip}{0pt}
    \item \textit{associativity}: $(x \cdot y) \cdot z = x \cdot (y \cdot z)$;
    \item \textit{identity}: there exists some $e \in G$ such that $e \cdot x = x = x \cdot e$ for all $x \in G$;
    \item \textit{inverses}: for every $x \in G$, there exists $x^{-1} \in G$  such that $x \cdot x^{-1} = e = x^{-1} \cdot x$.
\end{enumerate}
The operator $\cdot$ enforces a structure on $G$, or equivalently expresses an underlying symmetry.
For example, let $G$ be the rotational group of the equilateral triangle, in which the operator is a composition of rotations and the elements are rotations of the triangle by multiples of $120\degree$.
Notice that this group of rotations -- in which there are only three non-degenerate elements -- precisely describes the geometric symmetry of the triangle; and conversely, the symmetric structure of the triangle may be defined exactly as this group $G$.
It is important to note that two different groups can express the same symmetry (i.e., are \textit{isomorphic}; see Appendix~\hyperref[app:A.1]{A.1} for details).
For example, $G$ of our example is isomorphic to the group of integers modulo 3, with elements $0, 1, 2$ and a group operation of addition modulo 3 (e.g., $2 + 2 = 1$).

\subsubsection{Symmetry Functions} \label{sec:2.1.1}

Groups are a natural setting in which to study functions that use symmetry for \textit{radical generalization} (i.e., extrapolation far beyond the training distribution, requiring more than just surface correlations).
This hinges on two central concepts, \textit{invariance} and \textit{equivariance}, that refer to the extent to which a functional form respects a given symmetry.
For example, single-digit modulus arithmetic can be thought of as reflecting an invariance -- insofar as it is used for all places in multi-digit arithmetic -- while its application across multiple digits (i.e., over multiple scales) reflects equivariance.
We formalize these concepts in the general case here, and then consider their specific application to base arithmetic in Section~\ref{sec:2.1.2}.
To treat these formally, consider the case in which we wish to learn a function $f: X \to Y$, the domain ($X$) and co-domain ($Y$) of which have a common symmetry -- i.e., they are acted on by the same group $G$ (see Appendix~\hyperref[app:A.1]{A.1} for details).
$f$ is invariant with respect to $G$ if transformations by $G$ leave $f$ unchanged; that is, if $f(g \cdot x) = f(x)$ for all $x \in X$ and $g \in G$.
More generally, $f$ is equivariant if $f$ intertwines with transformations by $G$; that is, $f(g \cdot x) = g \cdot f(x)$ for all $x \in X$ and $g \in G$.\footnote{In an abuse of notation, here $\cdot$ refers to the action on both $X$ and $Y$.}
In this respect, invariance is a special case of equivariance in which the action of $G$ on $Y$ is trivial (i.e., $g \cdot y = y$ for all $g \in G, \, y \in Y$).
Both invariance and equivariance imply that $f$ respects the common symmetry of $X$ and $Y$; hence, we call such an $f$ a \textit{symmetry function}.

Equivariance is a crucial prerequisite for radical generalizability:
Consider the function $f$ that has been observed only on some small sub-domain $A \subset X$ sufficient to reveal the underlying symmetry (i.e., the action of $G$ on each $X$ and $Y$).
If $f$ is equivariant, then we may infer $f(y) = g \cdot f(x)$ for any $y = g \cdot x \notin A$.
Conversely, if $f$ is not equivariant, it becomes impossible (at least on the basis of group-theoretic structure alone) to broadly extrapolate from the sub-domain $A$ to its orbit $G \cdot A$ (see Appendix~\hyperref[app:A.1]{A.1} for details).
Therefore, radical generalization is possible if and only if the unknown function $f$ is equivariant to some known symmetry -- i.e., if $f$ is a symmetry function.
Accordingly, (i) all spaces relevant to radically generalizable function learning have symmetry; and (ii) all radically generalizable functions we wish to learn are symmetry functions.

\subsubsection{Group Extensions} \label{sec:2.1.2}

One final group theoretic construct relevant to our considerations
is that of a \textit{group extension}, which refers to the way that a group $A$ can be ``extended'' by another group $G$ to form the extension $E$; specifically, $E$ is a group in which $G$ and $A$ become embedded.
The use of clocks to represent time provides an instructive example, and one that is closely related to the case of base addition that we consider below: the clock ($E$) can be seen as a rotational group (the hour hand $A$, with 12 non-degenerate elements), extended to have an additional scale of structure -- the minute hand ($G$) -- using a similar symmetry (in that case with 60 non-degenerate elements).

Two extensions $E$ and $E'$ are \textit{equivalent} if, not only are $E$ and $E'$ isomorphic as groups, but $A$ and $G$ are embedded in ``the same way'' in $E$ and $E'$.
Just as we represent time using hours and minutes, we may seek to represent an extension $E$ on the set $A \times G$; doing so reduces to constructing a group operation on the set $A \times G$ that makes it an equivalent extension to $E$.
Here, \textit{group cohomology} -- a mathematical method for studying groups by chaining them into sequences -- provides a useful approach: in particular, we use the concepts of cocycles and coboundaries.
In the context of group extensions, a \textit{cocycle} is the part of the group operation on $A \times G$ (specifically, a function $f: G \times G \to A$) that makes the extension on $A \times G$ equivalent to $E$, whereas a \textit{coboundary} is the ``difference'' between the cocycles of two equivalent extensions both represented on $A \times G$.
In a clock, the cocycle is the function specifying that the hour hand increments when the minute hand crosses 60.
And, considering an alternative clock that counts counterclockwise in the minute hand, the coboundary is the function reversing the minutes (for minute $m$, $60 - m$) of a normal clock.

Morning (AM) vs. afternoon/evening (PM), days, and years all reflect repeated extensions of the same form of symmetry (a rotational group, differing only by the number of elements).
If applied properly, this repeated extension of a rotational group achieves equivariance, such that the system is synchronized across scales.
As such, in this paper we frequently refer to invariance and equivariance as properties of a repeated extension, insofar as: the symmetry invoked at each scale is invariant; and, if the extension is valid, its group operation is equivariant (i.e., associative).
For all of the above, see Appendix~\hyperref[app:A.2]{A.2} for a formal treatment.
Below, we will formulate base addition in these terms and, in so doing, create various scales of structure within the integers (in base 10, the 1's place, 10's place, 100's place, and so on).

\subsection{Base Addition as a Symmetry} \label{sec:2.2}

The integers are a paradigmatic symmetry group, that we denote by $\Z$: its elements are integers, and the group operation is addition ($+$; henceforth called \textit{integer addition}).
Humans exploit this for radical generalization through the use of base addition.
To see this, we can rewrite $+$ as a function $a: \Z \times \Z \to \Z$.
Now, suppose $a$ is known over finite $A \subset \Z \times \Z$ (e.g., examples of addition involving only two or three digits\footnote{Note that an integer in $\Z$ has no inherent structure such as digits -- rather, that is exactly what we will construct in its base representation -- but we refer to digits here in part for clarity (reflecting just how fundamental a base representation is to our understanding of integers) and to motivate the usefulness of a base representation.}), and we wish to learn $a$ over all of $\Z \times \Z$ (i.e., addition over numbers with any number of digits).\footnote{Our setup is reminiscent of Saul Kripke's skeptical response to Wittgenstein on rule-following: supposing $a$ is known only over finite $A$, how can we know that we are properly generalizing $a$ to $\Z \times \Z$, and not instead some alternative $a'$ which agrees with $a$ over $A$? We would argue that it follows from the construction of a base representation, which in turn relies on the axioms of group theory and cohomology. By making explicit what we assume, mathematical axioms address -- without necessarily offering a solution to -- the infinite regress problem posed by Kripke's skeptic.}
Given a minimally adequate size of $A$ (viz., a number of digits sufficient to exemplify all possible carries) then, insofar as $a$ is equivariant to translation by $\Z$ (implied by the associativity of addition; i.e., $(x + y) + z = x + (y + z)$ may be rewritten as $a(x + y, z) = x + a(y, z)$), the remaining $\Z \times \Z - A$ that can be computed is infinite.
That is, learning from a minimal set of examples $A$ (that reveal the underlying symmetry), is sufficient to support generalization over the entire domain $\Z \times \Z$ of the function $a$.

The foregoing shows how base addition can serve as a simple, well-characterized, and fundamental example of how discovering symmetry functions can support learning efficiency and radical generalization.
It is clear that humans exploit this when learning and carrying out base addition: we have developed an algorithmic kernel that exploits the symmetry of $\Z$ and can be applied recursively over $a$.
In this solution, the kernel itself is invariant (i.e., adding modulo base $b$ in each place), and the desired equivariance is implemented by ``extending'' that kernel via a recursive ``place-keeping'' function (i.e., successively carrying between digits).

More precisely, at the core of the solution is $\Z_b$: the finite group of integers modulo $b$, comprised of elements $0, 1, ..., b-1$ and the group operation \textit{addition modulo} $b$.
It is a foundational property of arithmetic that we may uniquely (up to leading zeros) represent any non-negative integer $n \in \Z_{\geq 0}$ as a sequence of such digits; that is, as a tuple $(n_k, n_{k-1} ..., n_1)$, where $n_j \in \Z_b$ is the $j$\textsuperscript{th} digit for $j \in [k]$.\footnote{Here, for simplicity, we limit our consideration to non-negative integers (for details on the group cohomological construction, see Appendix~\hyperref[app:A.2]{A.2}).  However, it is straightforward -- albeit not as elegant -- to include the negative integers, by including them as elements and modifying the group operation appropriately.}
We call such multi-digit tuples a number's \textit{base representation}.
This characteristic of arithmetic is reflected in the fact that, for decimal addition (i.e., base 10), we do not have unique symbols for integers higher than 9.
However, in order to faithfully reproduce the structure of $\Z$, it is necessary to construct an equivalent addition operator in the base representation (viz., a group operator which makes the base representation equivalent to the integers), which we call \textit{base addition}.
\citet{Isaksen2002} spells this out for the 2-digit case in base 10 (i.e., up to 99), and \citet{Segert2024} extends this to the general case of multi-digit numbers with any length.
Constructing a base addition operator amounts to defining a procedure for ``carrying'' an appropriate number to the next place depending on the pair of digits at the preceding place, which we call a \textit{carry function}.
This, in turn, can be formulated in terms of cocycles and group cohomology (introduced in Section~\ref{sec:2.1.2}).
As we elaborate below, doing so reveals a variety of carry functions for bases greater than 2, that we go on to characterize in subsequent sections.

\subsection{The Cohomological Construction of Base Addition} \label{sec:2.3}

Following \citet{Isaksen2002}, we construct base addition using the formalism of group cohomology.
Here, we present the construction in an intuitively accessible form; a more rigorous mathematical treatment is provided in Appendix~\hyperref[app:A.2]{A.2}.
The problem may be stated as follows.
Suppose we have two non-negative integers $n$ and $m$ in their base representations $(n_{k_n}, ..., n_1)$ and $(m_{k_m}, ..., m_1)$.
What is the sum $s = n + m$ in its base representation $(s_{k_s}, ..., s_1)$?\footnote{Here, we consider adding two non-negative integers, but note that if the base representation is properly constructed, this naturally extends to adding three or more non-negative integers using the associativity of addition.}

In arithmetic, humans are first taught to approach this problem sequentially, by adding the digits in each place: start with the least significant (rightmost) place; add the digits in that place modulo the base; and, if the sum in that place is greater than or equal to the base, carry 1 over to the addition of the digits in the next-most-significant (left-adjacent) place; then the apply the same algorithm to that next place, and so forth.
This can be formalized in terms of  the base representation of $s$ as:
\begin{align} \label{eq:1}
    \begin{split}
        s_j &= n_j + m_j + c_j, \\
        c_{j+1} &= \I_{n_j + m_j + c_j \geq b},
    \end{split}
\end{align}
where $c_j$ is the carry to the $j$\textsuperscript{th} digit (note, $c_1=0$).
We call this particular carry function (i.e., $c_{j+1} = \I_{n_j + m_j + c_j \geq b}$) the $\mathbf{1}$ carry function.
However, as we consider below, it is just one of several possible ways of carrying for bases greater than 2.
There are two important and separate factors to consider about Equation~\ref{eq:1}: validity and efficiency.
With respect to validity, this base addition formula is equivalent to integer addition (a formal explanation is provided in Appendix~\hyperref[app:A.2]{A.2}).
With respect to efficiency, this procedure -- a simple algorithmic kernel applied recursively (i.e., repeatedly over the pairs of digits at each place, in order of their increasing significance) -- is remarkably compact: it reduces learning addition in infinite $\Z$ to learning addition in finite $\Z_b$ and the $\mathbf{1}$ carry function.

\subsubsection{Carry Functions} \label{sec:2.3.1}

In addition to the $\mathbf{1}$ carry function, it is possible to identify other carry functions and evaluate both their validity and efficiency. This may be of interest for at least two reasons: (i) mathematical inquiry (what carry functions preserve the structure of the integers, and what might their associated base representations look like?); and (ii) learnability (e.g., where there is more than one carry function, to what extent do they vary in complexity and, consequently, the ease with which they can be learned?).

To consider different carry functions, we can generalize the procedure for base addition in Equation~\ref{eq:1} as follows.
First, consider the 2-digit case.
Given $f: \Z_b \times \Z_b \to \Z_b$, the base representation of $s$ is given simply by
\begin{align*}
    s_1 &= n_1 + m_1, \\
    s_2 &= n_2 + m_2 + f(n_1, m_1)
\end{align*}
where $f$ is the carry function.\footnote{Note that, in the 2-digit case and the general case (Equation~\ref{eq:2}), if two numbers differ in length, we zero-pad the shorter one from the left to match the length of the longer one.}
For an arbitrary number of digits, the carry function must take account of what has been carried previously; hence, the $j$\textsuperscript{th} digit of $s$ is given by
\begin{align} \label{eq:2}
    \begin{split}
        s_j &= n_j + m_j + c_j, \\
        c_{j+1} &= f(n_j, m_j) + f(n_j + m_j, c_j),
    \end{split}
\end{align}
where $f$ is again the carry function, and $c_1=0$ as in Equation~\ref{eq:1} (see Appendix~\hyperref[app:A.2]{A.2} for a derivation of Equation~\ref{eq:2}).

\subsubsection{Carry Function Validity and $k$-Equivariance} \label{sec:2.3.2}

For a carry function $f$ to be valid, it must preserve the structure of $\Z$.
There are two necessary and sufficient conditions for this to hold: (i) preservation of associativity, ensuring that the resulting base representation forms a group; and (ii) equivalence to integer addition, ensuring that the base representation is isomorphic to $\Z$ in particular.

Regarding associativity, a given carry function $f$ may exhibit this -- and hence the corresponding base addition may be equivariant -- only up to $k$-digit numbers, where $k \in \N$ can vary, and that we refer to as $k$-equivariance.
In the simplest 2-digit case, a carry function $f$ preserves associativity if and only if $f$ is a cocycle.
In general, we say that a carry function $f$ is \textit{$k$-equivariant} if it preserves associativity up to $k$ digits, and \textit{$\infty$-equivariant} if it does so for an arbitrary number of digits (see Appendix~\hyperref[app:A.2]{A.2} for details).
The notion of $k$-equivariance suggests one way to categorize carry functions; we consider others in the sections that follow, observing that these partly align with $k$-equivariance.

As with associativity, preservation of the equivalence of a carry function's corresponding base addition to integer addition may also be limited to $k$ digits.
Formally, base addition (as defined in Equation~\ref{eq:2}) is equivalent to integer addition up to $k$ digits if and only if $f$ is a $k$-equivariant cocycle differing from the $\mathbf{1}$ carry function by a coboundary.
For a more complete discussion of how to construct base addition with the formalism of group cohomology -- including the role of cocycles, coboundaries, and $k$-equivariance -- see Appendix~\hyperref[app:A.2]{A.2}.

\subsection{Classifying Carry Functions} \label{sec:2.4}

In addition to their range of equivariance, carry functions may be characterized in a number of other ways that are useful both for categorizing them mathematically, and for understanding how easily they can be learned.
Here, we focus on the identity of the values that are carried.
Specifically, carry functions can be divided into those for which the same integer value is always carried (\textit{Single Value} carry functions), and those for which different integers may be carried depending on the pair of digits (\textit{Multiple Value} carry functions).

\subsubsection{Single Value Carry Functions} \label{sec:2.4.1}

A carry function is designated as \textit{Single Value} if the same integer value is carried in all cases except when nothing (i.e., 0) is carried. 
The paradigmatic example is the $\mathbf{1}$ carry function introduced in Section~\ref{sec:2.3}; referencing Equation~\ref{eq:1} and Equation~\ref{eq:2}, its functional form is given by
\begin{equation} \label{eq:3}
    f_1(n, m) = \I_{n + m \geq b}.
\end{equation}
This produces the base representation of the integers  commonly taught in arithmetic.
However, other Single Value carry functions are also possible, that we refer to as \textit{alternative} Single Value carry functions.
These use a different carry value.
In general, any coprime to $b$ can serve as an alternative Single Value carry, where \textit{coprime} refers to any number $<b$ that is not a factor of $b$.
We refer to a coprime to $b$ as a \textit{unit} $u \in \Z_b$ of $b$, any of which can be used in a Single Value carry function, that we refer to as a $\mathbf{U}$ carry function.
This is because incrementing by a unit ($u$) covers all $b$ elements of $\Z_b$ in just $b$ increments.
However, each does so in a different order.  

To see this, first consider the $\mathbf{1}$ carry function (noting that 1 is a unit for any $b$): incrementing  by 1 always covers all $b$ elements of $\Z_b$ in just $b$ increments.
In this case, the increments follow the natural order of the integers in $\Z_b$ (see the first row of the example below).
 
In contrast, consider counting up to two digits in base $b=3$.
Since here, 2 is also unit $u$ of $b$, it can be used in a $\mathbf{2}$ carry function, as shown in the example below.
As a reference, the first row shows counting using the $\mathbf{1}$ carry function, in which each increment is by $(0, 1)$; the second row shows the $\mathbf{2}$ carry function, in which each increment is by $(0, 2)$:
\begin{align*}
    \text{$\mathbf{1}$ carry function:} \quad
    (0, 0),\, (0, 1),\, (0, 2),\, (1, 0),\, (1, 1),\, (1, 2),\, (2, 0),\, (2, 1),\, (2, 2). \\
    \text{$\mathbf{2}$ carry function:} \quad
    (0, 0),\, (0, 2),\, (0, 1),\, (2, 0),\, (2, 2),\, (2, 1),\, (1, 0),\, (1, 2),\, (1, 1).
\end{align*}
Note that for both carry functions, all values of $\Z_b$ are covered exactly once in the least significant (rightmost) place before the second (leftmost) place is used; they are simply covered in a different order.
Figures \ref{fig:1} and \ref{fig:2} show graphic representations of the carry functions for $b=3$ and 4.

In general, every unit $u$ can be used for an alternative Single Value carry function -- that is associated with a different and equally valid ordering of $\Z_b$  --  in which either 0 or u is carried for all pairs of digits.
Importantly, all of the Single Value carry functions are $\infty$-equivariant (see Section~\ref{sec:2.3.2} and Appendix~\hyperref[app:A.2]{A.2}).

\subsubsection{Multiple Value Carry Functions} \label{sec:2.4.2}

Finally, we note that for bases beyond $b=2$, there are carry functions that do not always carry the same digit, and that cannot be reduced to the $\mathbf{1}$ carry function in the way discussed above.
We refer to these as {Multiple Value} carry functions, which comprise a more heterogeneous group than Single Value carry functions.
To our knowledge, there is no full, formal characterization of the relative distribution of Multiple Value versus Single Value carry functions for a given base $b$.
However, to gain some insight into how this scales with $b$, we have calculated that, for 2-digit addition, there are $b^{b-2}$ equivalent carry functions (this can be shown by fixing a cocycle and counting the possible coboundaries; see Appendix~\hyperref[app:A.2]{A.2}).
Of these, $\varphi(b)$ are Single Value carry functions, and the remaining $b^{b-2} - \varphi(b)$ are Multiple Value carry functions.
$\varphi(b) \leq b$, so the fraction of Multiple Value carry functions is at least $b^{b-2} - b$.
Thus, the fraction of Single Value carry functions vanishes for large $b$.

In Section~\ref{sec:3}) we observe that there is a subset of Multiple Value carry functions that are notably lower in complexity than the remainder.
We refer to these as \textit{Low Dimensional Multiple Value carry functions}, that we characterize in  Section~\ref{sec:3}), and return to in Section~\ref{sec:5.1} where we consider their base representations and the extent to which they exhibit $\infty$-equivariance.
\section{Quantitative Measures of Carry Functions} \label{sec:3}

As noted above, the number of carry functions grows rapidly with $b$, and these are heterogeneous both with respect to the nature of their associated carries and extent of their equivariance.
Here, we consider quantitative measures that characterize of carry functions, as means of categorizing them, and for use in considering the efficiency with which they can be learned.
To do so, it is useful to represent carry functions in matrix form, as ``carry tables''.

\subsection{Carry Tables} \label{sec:3.1}

For a given base $b$, we can represent any carry function $f: \Z_b \times \Z_b \to \Z_b$ as a matrix $F \in (\Z_b)^{b \times b}$, where $F_{n,m} = f(n, m)$.
The matrix $F$ is referred to as a \textit{carry table}, that provides a graphical representation of the carry function $f$ in which each element designates the value that is carried (indicated by color) depending on the pair of digits indexed by the corresponding row and column.
Figure~\ref{fig:1} shows the 16 carry tables in base 4 (see Appendix~\hyperref[app:B.1]{B.1} for the carry tables in base 5).
The Single Value carry functions are outlined in blue (the $\mathbf{1}$ carry function in the top left and the $\mathbf{3}$ carry function, bottom right), the Low Dimensional Multiple Value carry functions in orange, and the Other Multiple Value carry functions in grey.
A carry table's entries are indexed by $\{0, 1, ..., b-1\}$ from left to right, top to bottom.
Thus, with this indexing, the $\mathbf{1}$ carry function is $F_{n,m}=0$  if $n + m < b$ or 1 if $n+m$  $\geq b$ (in agreement with Equation~\ref{eq:3}).
While the other carry functions are all valid (i.e., producing equivalent base additions to that of the $\mathbf{1}$ carry function, at least for 2-digit numbers), their structure differs, as shown in the figure, and quantified in Section \ref{sec:3}.

\begin{figure}[h]
    \centering
    \includegraphics[width=0.7\textwidth]{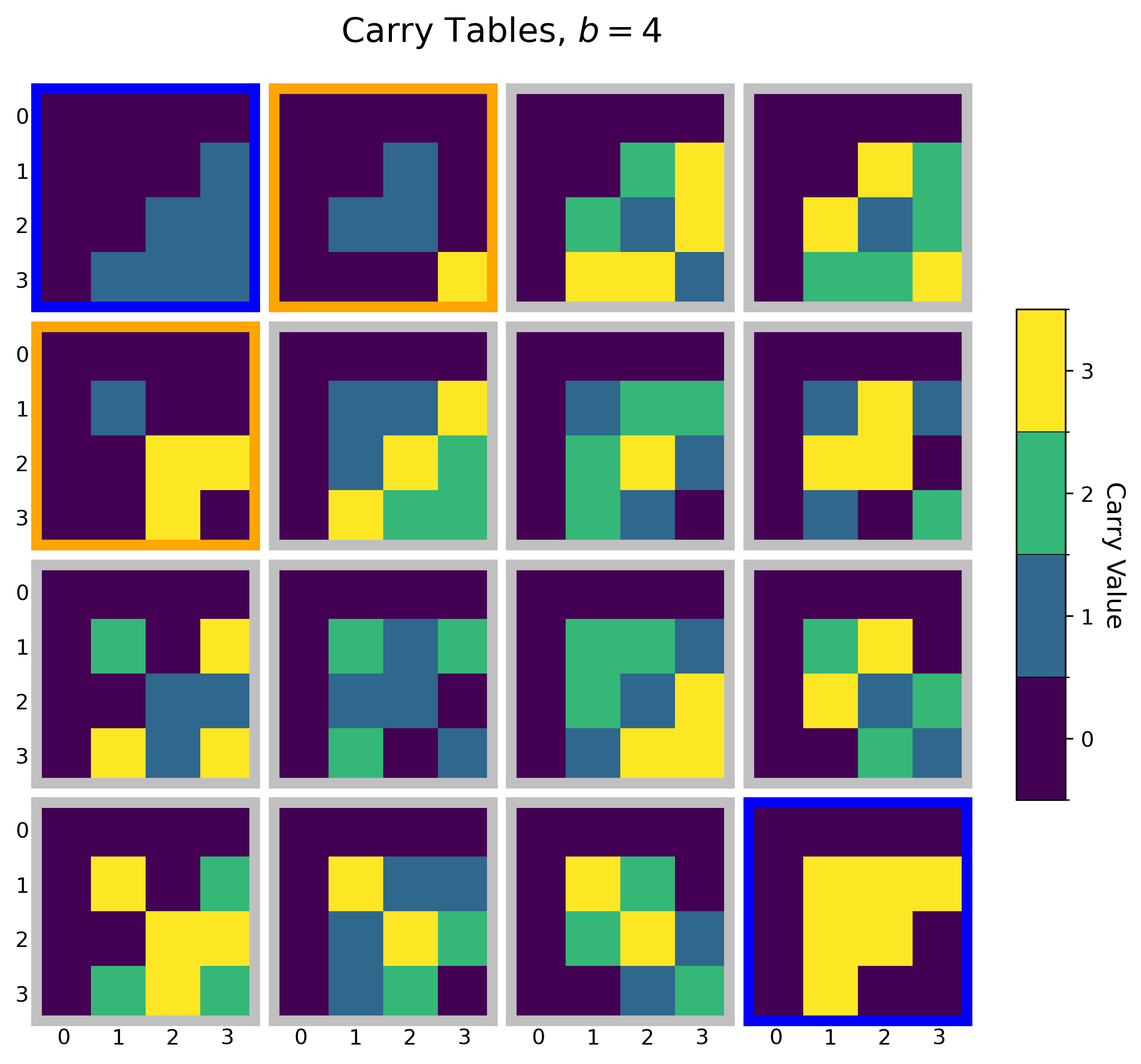}
    \caption{
    The 16 carry tables in base 4.
    The Single Value carry functions are outlined in blue (the $\mathbf{1}$ carry function in the top left and the $\mathbf{3}$ carry function, bottom right), the Low Dimensional Multiple Value carry functions in orange, and the Other Multiple Value carry functions in grey (see Section~\ref{sec:2.4}).
    Each table's entries are indexed by $\{0, 1, ..., b-1\}$ from left to right, top to bottom; color indicates the value that is carried (see legend at right).
    }
    \label{fig:1}
\end{figure}

As noted earlier, however, some carry functions are only finitely equivariant.
For example, while the $\mathbf{1}$ carry function (top left of Figure~\ref{fig:1}) is $\infty$-equivariant (i.e., associative for triplets of any length), the Multiple Value carry function in the top right of Figure~\ref{fig:1} is only 2-equivariant, and thus valid for only 2-digit numbers (i.e., not associative for all 3-digit triplets; e.g., $((0, 0, 1) + (0, 0, 2)) + (0, 0, 3) = (3, 2, 2)$ whereas $(0, 0, 1) + ((0, 0, 2) + (0, 0, 3)) = (0, 2, 2)$.
While $k$-equivariant carry functions are, by definition, limited in their extension (i.e., the scale of their resulting base addition's equivariance), they may be of interest in settings where addition must learned, as solutions that may be discovered when the training set is constrained to numbers of length $\leq k$.
We return to this consideration in Section~\ref{sec:3.4}.

\begin{figure}[h]
    \centering
    \includegraphics[width=0.7\textwidth]{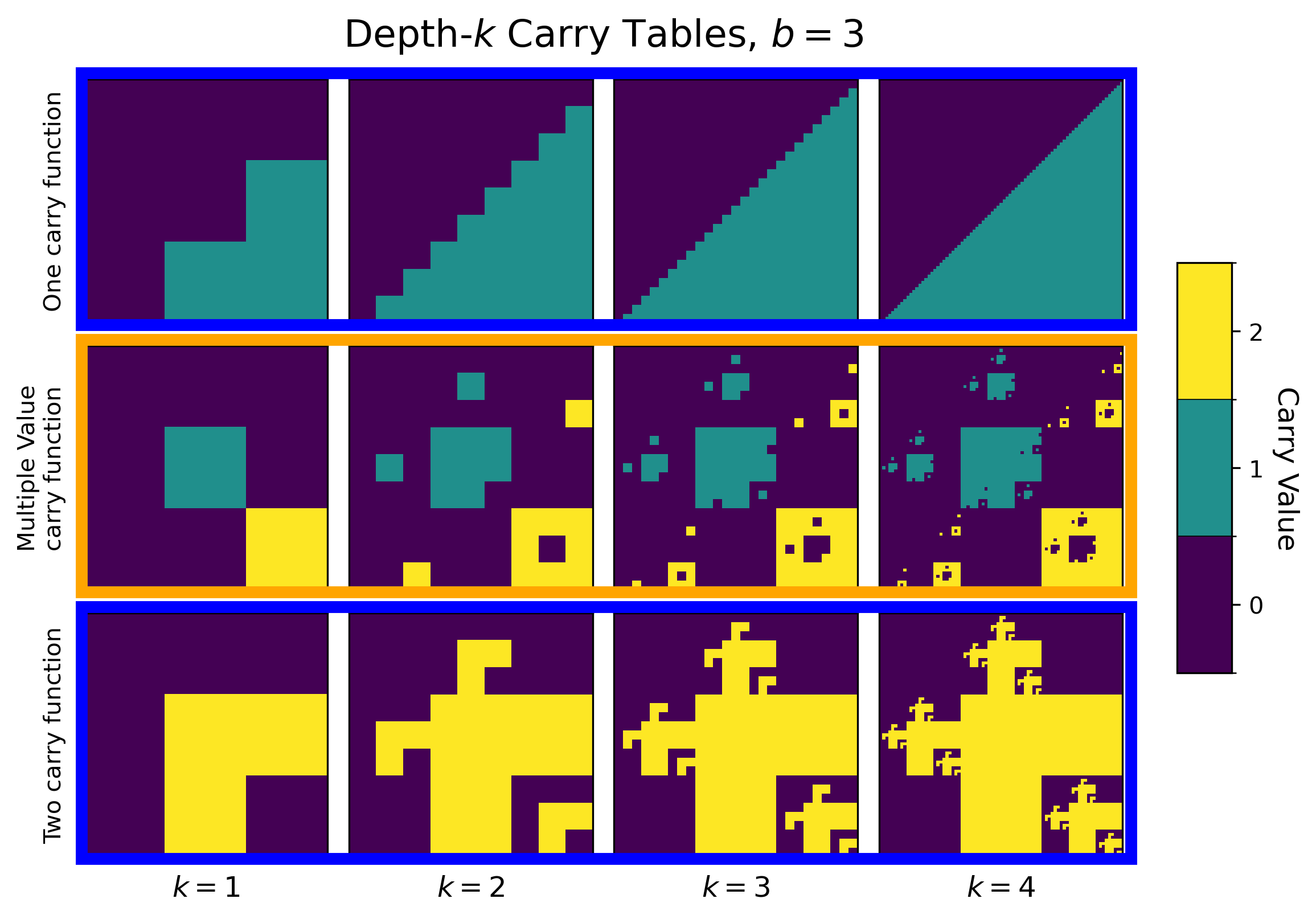}
    \caption{
    Structure at increasing depths of the three carry tables for $b=3$.
    The panels in each row correspond to different carry functions, and those in columns show the depth-$k$ carry tables for each carry function.
    As in Figure~\ref{fig:1}, the Single Value carry functions are outlined in blue (the $\mathbf{1}$ carry function in the top row and the $\mathbf{2}$ carry function in the bottom row) and the Low Dimensional Multiple Value carry function is outlined in orange (there are no Other Multiple Value carry functions for base 3).
    For depths $k=1$ to 4 (columns), each depth-$k$ carry table $F_k$ (see Section~\ref{sec:3.1}) is indexed from left to right, top to bottom by $(n_k, ..., n_1)$ in the usual lexicographical ordering, and colors indicate the integer value carried for each pair of digits added.
    Note that the scale of each axis grows by a factor of $b$ as $k$ is increased.
    }
    \label{fig:2}
\end{figure}

To more precisely characterize the equivariance of carry functions, we define a \textit{depth-k carry table} as $F_k \in (\Z_b)^{b^k \times b^k}$.
The entries of $F_k$ correspond to what is carried from the $k$\textsuperscript{th} digit to the $k+1$\textsuperscript{th} digit, such that, for $n = (n_k, ..., n_1)$ and $m = (m_k, ..., m_1)$ in the usual lexicographical ordering, $(F_k)_{n, m} = c_{k+1}$, for $c_{k+1}$ as given in Equation~\ref{eq:2}.
We refer to $k$ as the \textit{depth} of the table.
Figure~\ref{fig:2} shows depth-1, depth-2, depth-3 and depth-4 carry tables for base 3.
For example, the table in the top, second from left panel of Figure~\ref{fig:2} shows the depth-2 carry table for the $\mathbf{1}$ carry function, confirming that at depth 2 a 1 is carried from the second to third digit for sums of (1, 0, 0) or more.
Note that this is true for all four depths shown, consistent with the fact that the $\mathbf{1}$ carry function is $\infty$-equivariant.
From Figure~\ref{fig:2} it is clear that, although all three carries are at least 4-equivariant, the $\mathbf{1}$ carry function has considerably simpler structure (is more compact) than the others, which becomes increasingly apparent at greater depths.
We consider different ways of quantifying this in the sections that follow.

\subsection{Fractal Dimension} \label{sec:3.2}

It is apparent that all of the depth-$k$ carry tables shown in Figure~\ref{fig:2} have at least partially fractal structure; that is, repeated, self-similar structure at progressively finer scales. 
This comports with the idea that a $k$-equivariant carry function has a self-similar structure up to some $k$ digits (i.e., scales).

These observations suggest that fractal dimension may be a natural way to characterize the equivariance of carry functions, and to quantify their complexity.
For example, in the simplest case, as $k \to \infty$, the carry table for the $\mathbf{1}$ carry function converges to a triangle (see top row of Figure~\ref{fig:2}), whereas  other carry functions exhibit more complex fractal structure.

Generally, definitions of dimension in fractal geometry take a measurement $N_\epsilon$ at scale $\epsilon$ (which ignores granularity at scales $< \epsilon$), and observe how the value changes as $\epsilon \to 0$.
Though the most widely-used fractal dimension is the Hausdorff dimension, here we use the simpler box-counting dimension, given its natural application to carry tables.
Letting $N_\epsilon$ be the number of boxes of side-length $\epsilon$ required to cover the fractal, the box-counting dimension $d_{\text{box}}$ is defined as
\begin{equation} \label{eq:4}
    d_{\text{box}} = \lim_{\epsilon \to 0} \frac{\log(N_\epsilon)}{\log(1/\epsilon)}.
\end{equation}
Notice that Equation~\ref{eq:4} can be re-written as $N_\epsilon \approx C (1/\epsilon)^{d_{\text{box}}}$ for constant $C$, which reduces to the usual notion of dimension for integral $d_{\text{box}}$ (see \cite{Falconer2014} for further discussion).

For present purposes, we measure the box-counting dimension of the border of the depth-$k$ carry table in the limit $k \to \infty$; indeed, a table's complexity might correspond to the boundary between distinct carries (comparatively, the box-counting dimension of the interior corresponds to the frequency of carrying in the limit $k \to \infty$; see Section~\ref{sec:3.3}).
To avoid over-counting, we define the border as those entries $(F_k)_{n,m}$ with different values than their left and upper neighbors.
Box-counting dimension is natural for the case of carry tables because the ``resolution'' of the carry table at a given depth $k$ is exactly boxes of width $\epsilon=1/b^k$ (see Figure~\ref{fig:2}).
Therefore, it suffices to count the number of entries on the border to measure $N_\epsilon$ at depth $k$, and then compute Equation~\ref{eq:4} by setting $\epsilon=1/b^k$.
For example, for the $\mathbf{1}$ carry function at depth 2 (top, second from left of Figure~\ref{fig:2}), we have $N_\epsilon = 8$ and $\epsilon = 1 / 3^2 = 1/9$, giving an estimated $d_{\text{box}} \approx 0.946$.

Importantly, as the conventional ordering of $\Z_b$ is just one of the valid orderings generated by its alternative Single Value carry functions (see Section~\ref{sec:2.4.1}), we consider the minimal box-counting dimension across all such orderings of the depth-$k$ carry table indices.
Figure~\ref{fig:3}A shows the estimated box-counting dimensions for bases $b = 3$, 4, 5 at depths $k=1$ to 4.

\subsection{Frequency of Carrying} \label{sec:3.3}

Another way to quantify differences between carry functions, that complements its fractal dimension, is how often a carry must be made.
For the carry at digit $k$, this reduces to the fraction of non-zero numbers in the depth-$k$ carry table.
To measure the overall frequency of carrying,
it suffices to average the frequency of carrying at each digit.
Figure~\ref{fig:3}B shows the overall frequency of carrying for bases $b = 3$, 4, 5 at depths $k = 1$ to 4.

\subsection{Associativity Fraction} \label{sec:3.4}

\begin{figure}[h]
    \centering
    \includegraphics[width=\textwidth]{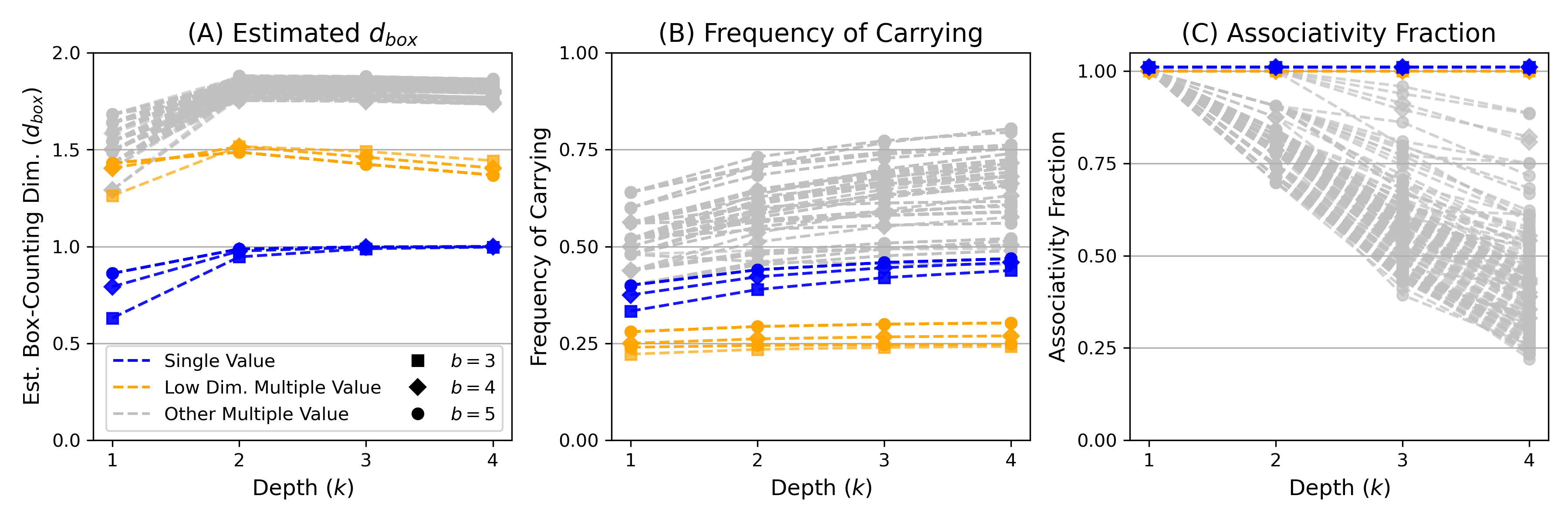}
    \caption{
    Quantitative characterization of different carry functions for bases 3 to 5 and depths 1 to 4, using three different measures: (A) estimated box-counting dimension of the fractal depth-$k$ table border; (B) frequency of carrying; and (C) associativity fraction.
    Each is shown for bases $b=3$ (as squares), 4 (diamonds), and 5 (circles) at depths $k=1$ to 4. 
    For each quantitative measure, as the depth increases the carry functions separate into the classes of the Single Value carry functions and Multiple Value carry functions.
    Furthermore, the Low Dimensional Multiple Value carry functions separate from the Single Value and other Multiple Value carry functions according to frequency of carrying as well as box-counting dimension.
    Note that, for associativity fraction, the Low Dimensional Multiple Value carry functions exactly underlie the Single Value carry functions (i.e., they all perfectly preserve associativity up to depth 4, and hence are 5-equivariant).
    }
    \label{fig:3}
\end{figure}

Finally, we quantify carry functions according to the validity of their resulting base addition, measured by their associativity (see Section~\ref{sec:2.1}).
Specifically, we consider the fraction of triplets for which a carry function's resulting addition is associative, up to varying depths.
That is, for each depth $k$, we sample triplets $(n, m, p)$ of length $k+1$ and measure the fraction that satisfy $(n + m) + p = n + (m + p)$, using the addition procedure of Equation~\ref{eq:2}.
Note that this measure is closely related to $k$-equivariance (see Section~\ref{sec:2.3}), since a $k$-equivariant carry function has an associativity fraction of 1 up to depth $k-1$.
Figure~\ref{fig:3}C shows the associativity fraction of carry functions for bases $b = 3$, 4, 5 at depths $k=1$ to 4.

Figure~\ref{fig:3} shows that each quantitative measure reflects the categorizations considered in Section~\ref{sec:2.4} (a correlation matrix comparing the measures is provided in Appendix~\hyperref[app:B.2]{B.2}).
First, at better estimates of the box-counting dimension (in particular, at depth $k=4$), the Single Value carry functions converge to a dimension of 1, whereas the Multiple Value carry functions all maintain dimension $>1.25$.
Among the Multiple Value carry functions, a subset converge to a dimension $>1.25$ but $<1.5$, corresponding to the Low Dimensional Multiple Value carry functions introduced in Section~\ref{sec:2.4.2}.
The Low Dimensional Multiple Value carry functions can also be distinguished from the other Multiple Value carry functions in the other quantitative measures: they have a lower frequency of carrying than not only the other Multiple Value carry functions, but also the Single Value carry functions; and they have associativity fraction of 1 up to at least depth $k=4$.
In the section that follows, we consider how these factors influence the efficiency with which carry tables can be learned by a neural network.
\section{Neural Network Simulations} \label{sec:4}

As noted in Section~\ref{sec:1}, humans exhibit a remarkable ability for radical generalization, as exemplified by their ability to carry out base arithmetic algorithmically, over an arbitrary number of digits.
This suggests that its implementation in the brain must respect the symmetry properties of base addition considered in this paper.
Here, we explore how the factors discussed in the previous sections, and their interaction with training procedures, may impact how neural networks discover the symmetries underlying base addition through learning. 
We begin by formulating the base addition problem in a format suitable for training a neural network, and describe the network architecture used for training and testing.
We then consider how the embedding of integers, training curriculum, and characteristics of various  carry functions impact learning and generalization.
Code for the network architecture and simulations using it are freely available at \href{https://github.com/cutterdawes/BaseAddition}{\texttt{https://github.com/cutterdawes/BaseAddition}}.

\subsection{Model Representations, Architecture, and Procedures} \label{sec:4.1}

\subsubsection{Stimulus Representations} \label{sec:4.1.1}

We formulated the addition problem for $k$-digit non-negative integers in base $b \in \N$, by constructing two multi-digit numbers,  $n = (n_k, ..., n_1)$ and $m = (m_k, ..., m_1)$, which were composed of lists of $b$ dimensional tensors representing each digit, ordered left to right from most significant ($_k$) to least significant ($_1$).
We considered two forms of digit representation (or \textit{embedding}): (i) a purely \textit{symbolic} embedding, in which each digit was represented as a different one-hot tensor, so that all were orthogonal to one another; and (ii) a \textit{semantic} embedding, in which numbers closer in ordinal value were more correlated with (similar to) one another than ones further apart (see Section \ref{sec:4.2}).
The two numbers $n$ and $m$ for a given problem could differ in length, in which case the shorter one was zero-padded from the left to match the length of the longer one.

\subsubsection{Network Architecture} \label{sec:4.1.2}

We were specifically interested in how a neural network can solve arithmetic problems when constrained to do so using serial processing, comparable to how humans do so.
We assume that this imposes a strong inductive bias favoring discovery of the symmetry properties of carry functions considered in the previous sections, advantaged by weight sharing over time inherent to recurrent neural networks.
Accordingly, the model was comprised of a single-layer GRU \citep[input dim. $b$, hidden dim. $b$;][]{Cho2014} followed by a single linear layer (input dim. $b$, output dim. $b$) used for decoding.

For all experiments, we also used a model with a single-layer LSTM \citep[input dim. $b$, hidden dim. $b$;][]{Hochreiter1997} in place of a GRU; refer to Appendix~\hyperref[app:C.1]{C.1} for details.

\subsubsection{Problem Format} \label{sec:4.1.3}

There are two distinct ways in which an addition problem can be presented.
In the most straightforward, all the digits of one of the numbers are presented first, in standard (most significant to least significant) order, followed by all the digits of the other number, after which the agent is expected to produce the result, also with digits ordered from most significant to least significant.
While this format is compact and familiar, it is also extremely difficult to learn, in part because it requires keeping track of and reordering the digits at encoding and response \citep{Segert2024}.\footnote{Hence, neural network solutions to this format may require the use of some form of external memory.}
While this is an important capability, our focus here was on how the arithmetic computations themselves are performed.
This format is also not the way humans learn to perform multi-digit arithmetic (perhaps for a similar reason).
Rather, they do so in what we refer to as an \textit{interleaved} format, that we used in our simulations.

In the interleaved format, one digit from each number is presented, beginning with the least significant; the numbers are added; the result is generated, and then the next-most significant set of digits is presented, etc.
That is, the input sequence $x$ is
\begin{equation*}
    x = (n_1, m_1, *, n_2, m_2, *, ..., n_k, m_k, *),
\end{equation*}
with special tokens denoting the (heldout) digits of the answer sequence $s = (s_k, ..., s_1)$.
In human schooling, this format is implemented by placing one full number above the other, right-aligned by the least significant digit, thus allowing addition of the numbers in each column, from left to right.
The problem then reduces to performing modular addition for the numbers in each column using $b$ as the modulus, generating the result, and carrying appropriately to the next column (i.e., implementing a carry function).
We simulated this procedure for the neural network by presenting it with the digits from each place, one pair at a time, in order from least to most significant, and requiring the network to generate the resulting digit for each place.
Accordingly, it had to learn both the base-appropriate modular addition, encode the appropriate carry in memory, and then include that in the next addition operation.
We trained it using the correct responses for different carry functions, and examined its ability to learn these as a function of the characteristics of carry functions considered in the previous section.

\subsubsection{Training and Testing Procedures} \label{sec:4.1.4}

Separate networks were trained to add 3-digit numbers using each possible carry function for bases $b=3$ to 5.
The loss was computed as the cross-entropy between the output logits in the response layer and the answer sequence determined by the corresponding carry function.
Each network was trained for 2500 epochs over all 3-digit tuples $(n, m)$, with a batch size of 32 and a learning rate of 0.05 using the Adam optimizer.
This was done for 10 initializations of each network.
To track learning and test generalization, we evaluated each network's performance on both 3-digit and 6-digit numbers every 10 epochs throughout training.

\subsection{Results} \label{sec:4.2}

\begin{figure}[h]
    \centering
    \includegraphics[width=0.9\textwidth]{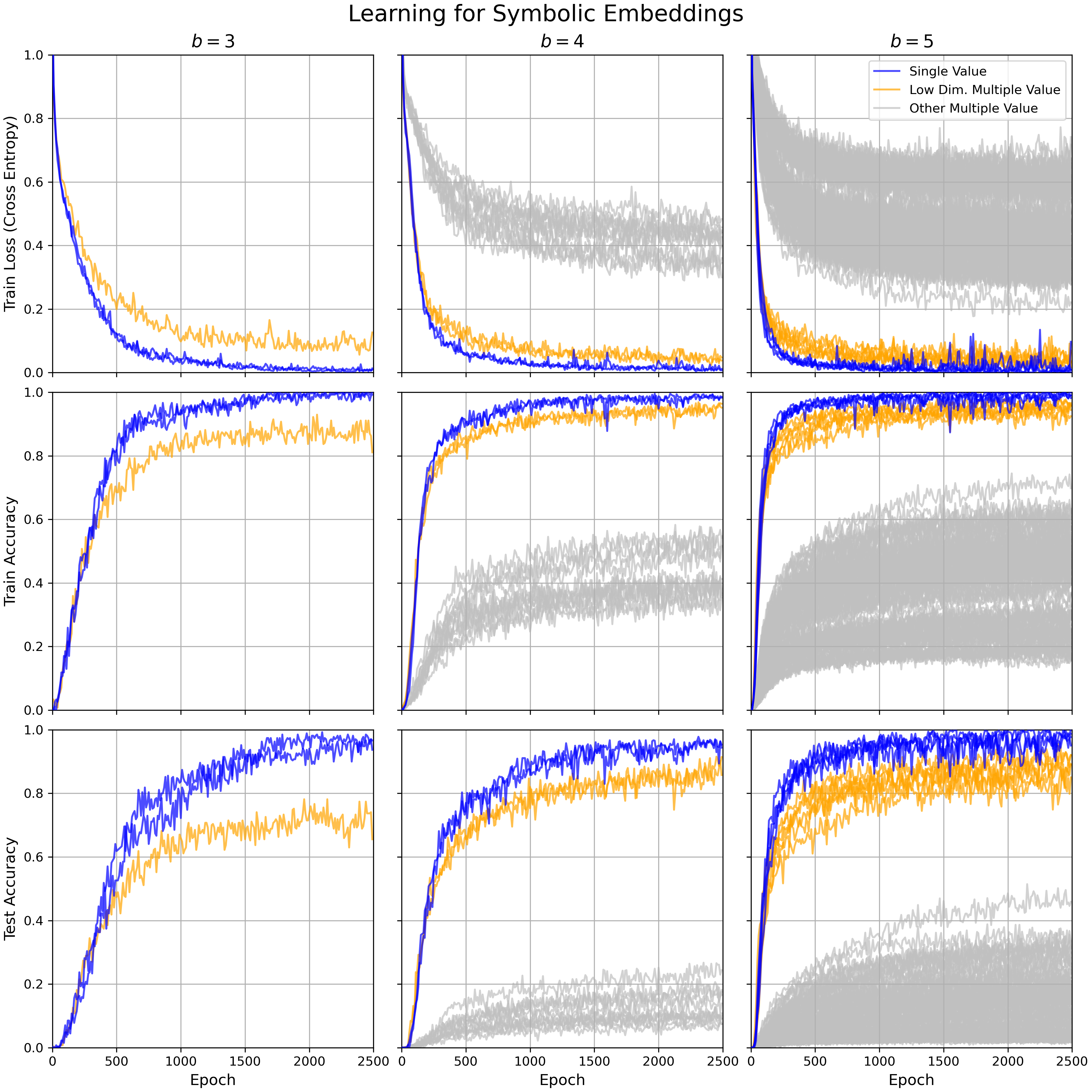}
    \caption{
    Learning for symbolic (one-hot) embeddings of digits.
    Performance over the course of training (averaged over 10 runs of each model implementation) for addition on different carry functions for bases $b=3$ to 5.
    }
    \label{fig:4}
\end{figure}

Figure~\ref{fig:4} shows the training loss as well as training and testing accuracy for bases 3 to 5, averaged for 10 runs of each configuration as noted above.
These results indicate that the model learned the Single Value carry functions and Low Dimension Multiple Value carry functions significantly more effectively than the other Multiple Value carry functions.

\subsubsection{Embedding} \label{sec:4.2.1}

The results shown in Figure~\ref{fig:4} are for symbolic (one-hot) embeddings of digits, in which there are no meaningful representational differences among the Single Value carry functions (see Section \ref{sec:2.4.1}).
This reflects the treatment of digits as symbolic labels, without any meaningful ordering.
However, whereas numbers are symbolic in one sense (that is, they can be used to refer to the quantify of any entity), they also have ordinal relationships, that humans learn prior to arithmetic.
To investigate how such information impacts the learning of different carry functions, we tested semantic embeddings of digits that reflected their ordinal relationships.
We did so by ordering one-hots to align with the ordinal value of each digit, and then convolving each with a Gaussian envelope mean-centered on its one-hot value (e.g., for $b=5$, the digit 1, formerly represented as the one-hot vector (0, 1, 0, 0, 0), became (0.2, 0.5, 0.2, 0.05, 0.05)).
Accordingly, each representation was now partially correlated with its immediate neighbors, and progressively less so with more distant ones.
Furthermore, because $\Z_b$ is cyclic (fundamental to its structure), we applied the convolution in circular form, so that so that 0 was equidistant to 1 and $b-1$.
We encoded alternative orderings of $\Z_b$ similarly (see Section~\ref{sec:2.4.1}).\footnote{Note that, due to the cyclic structure of $\Z_b$, a unit and its inverse produce the same ordering: in $b=5$ for example, the unit 4 (inverse of 1) produces the ordering (0, 4, 3, 2, 1), which produces the same semantic embeddings as (0, 1, 2, 3, 4).}
For example, corresponding to the ordering that results from incrementing by the unit 2 for $b=5$ (0, 2, 4, 1, 3), we encoded 1 as (0.05, 0.5, 0.05, 0.2, 0.2).
To examine how different orderings impact the learning of Single Value carry functions, we trained different implementations of the same model with the Gaussian-convolved one-hots corresponding to each ordering for that base (for 10 initializations of each network).

\begin{figure}[h]
    \centering
    \includegraphics[width=0.6\textwidth]{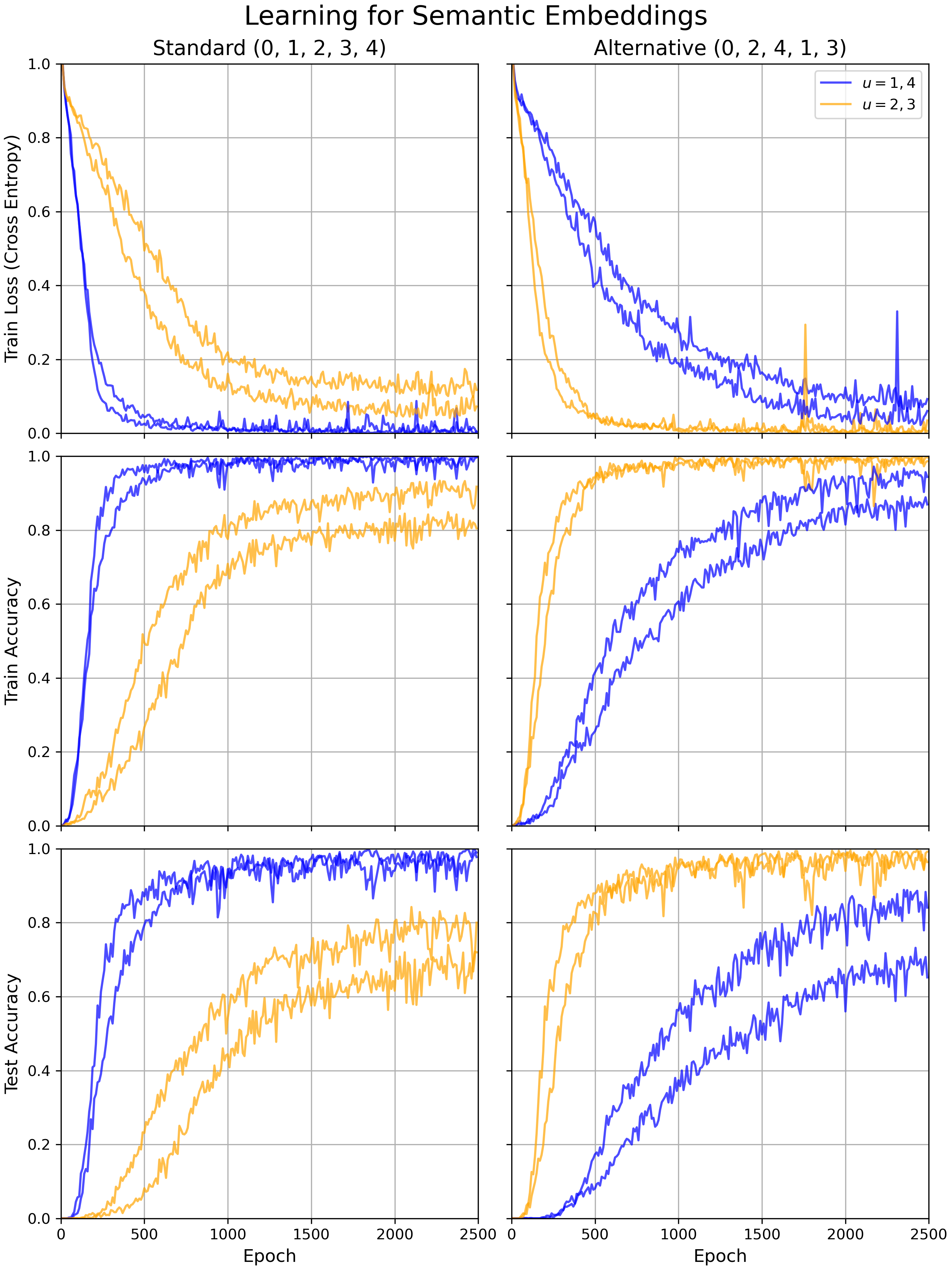}
    \caption{
    Learning for semantic embeddings of digits.
    Performance over the course of training on Single Value carry functions with order-encoded inputs in the two non-degenerate orderings (0, 1, 2, 3, 4) and (0, 2, 4, 1, 3) for $b=5$ (see Section~\ref{sec:2.4.1} for how Single Value carry functions correspond to orderings of $\Z_b$).
    }
    \label{fig:5}
\end{figure}

Figure~\ref{fig:5} shows the training results for semantic embeddings using $b=5$, for which there are four Single Value carry functions.
These separate clearly into two groups, each corresponding to an ordering of $\Z_b$: one that includes the standard $\mathbf{1}$ carry function (involving incrementing by 1; so 0, 1, 2, 3, 4) and its inverse, the $\mathbf{4}$ carry function; and a second set involving the $\mathbf{2}$ carry function (involving incrementing by 2; so 0, 2, 4, 1, 3) and its inverse, the $\mathbf{3}$ carry function.
Interestingly, for the semantic embeddings of each ordering, the model learned the set of Single Value carry functions corresponding to that ordering more quickly, indicating that it is comparatively easier to learn a carry function that reflects the ordinal structure of the digits (encoded in their embeddings) -- despite no difference in structure according to the quantitative measures of Section~\ref{sec:3}.

\subsubsection{Generalization} \label{sec:4.2.2}

The previous experiments tested generalization on 6-digit numbers, compared to the 3-digit numbers seen in training.
To probe generalization further out of domain, we tested accuracy on numbers up to 10 digits.
That is, for each carry function, we trained the model on 3-digit numbers and then tested on $k$-digit numbers for each $k \in [3:10]$, averaged over 10 training/testing runs.

\begin{figure}[h]
    \centering
    \includegraphics[width=\textwidth]{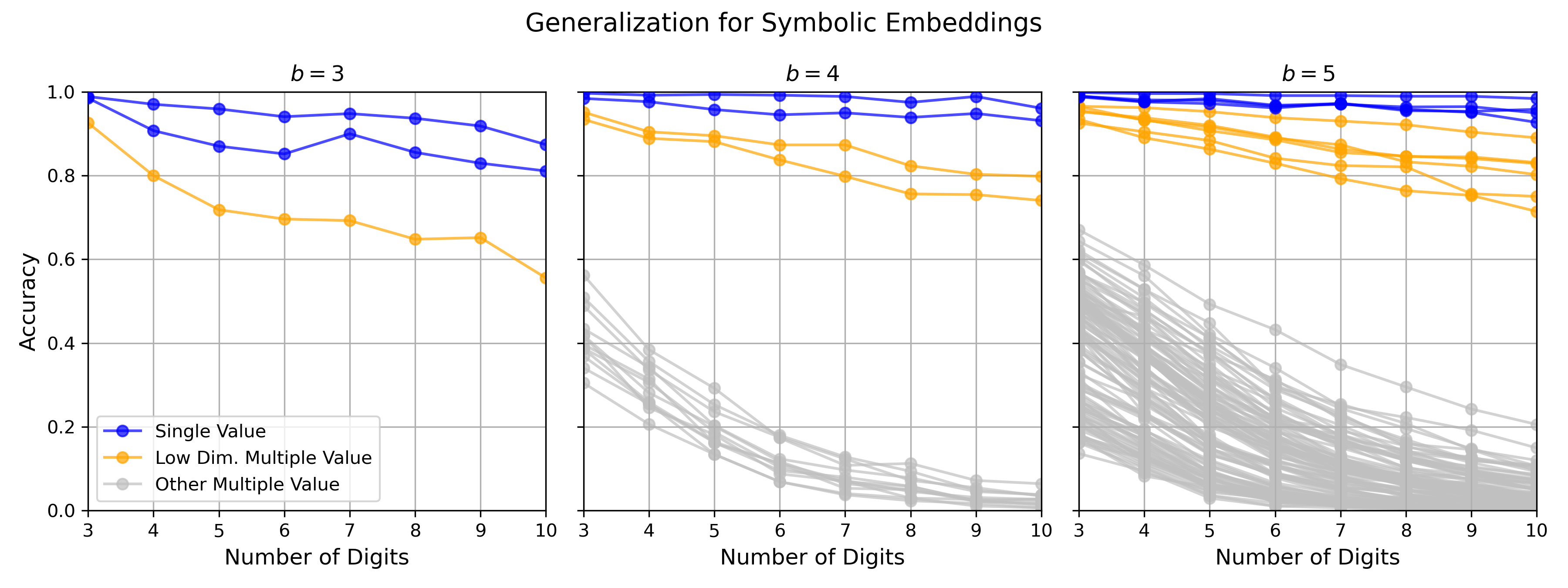}
    \caption{
    Out-of-domain generalization for symbolic embeddings of digits.
    For each carry function in bases $b=3$ to 5, accuracy was tested on $k$-digit numbers for each $k \in [3:10]$ after training on 3-digit numbers (averaged over 10 training/testing runs).
    }
    \label{fig:6}
\end{figure}

In particular, we did this for symbolic embeddings of digits in bases 3 to 5, shown in Figure~\ref{fig:6}.
Across all lengths and bases, the same separation between Single Value, Low Dimensional Multiple Value, and Other Multiple Value carry functions that was evident in the previous experiments persisted, and even became more apparent as the number of digits increased.
This corroborated that: (i) when learning a carry function of sufficient symmetry (viz., Single Value), a very small model can generalize far out of domain; and (ii) the symmetry of the learned carry function is increasingly important to performance as we move further from the training distribution.

Figure~\ref{fig:6} also reveals a somewhat surprising trend: in each class of carry functions (Single Value, Low Dimensional Multiple Value, Other Multiple Value), out-of-domain performance increases as the base increases from 3 to 5.
We discuss possible reasons for this trend in Section~\ref{sec:5.2.2}.
However, not surprisingly, carry function structure also has an influence on learning, as we consider next.

\subsubsection{Carry Function Structure} \label{sec:4.2.3}

\begin{figure}
    \centering
    \includegraphics[width=\textwidth]{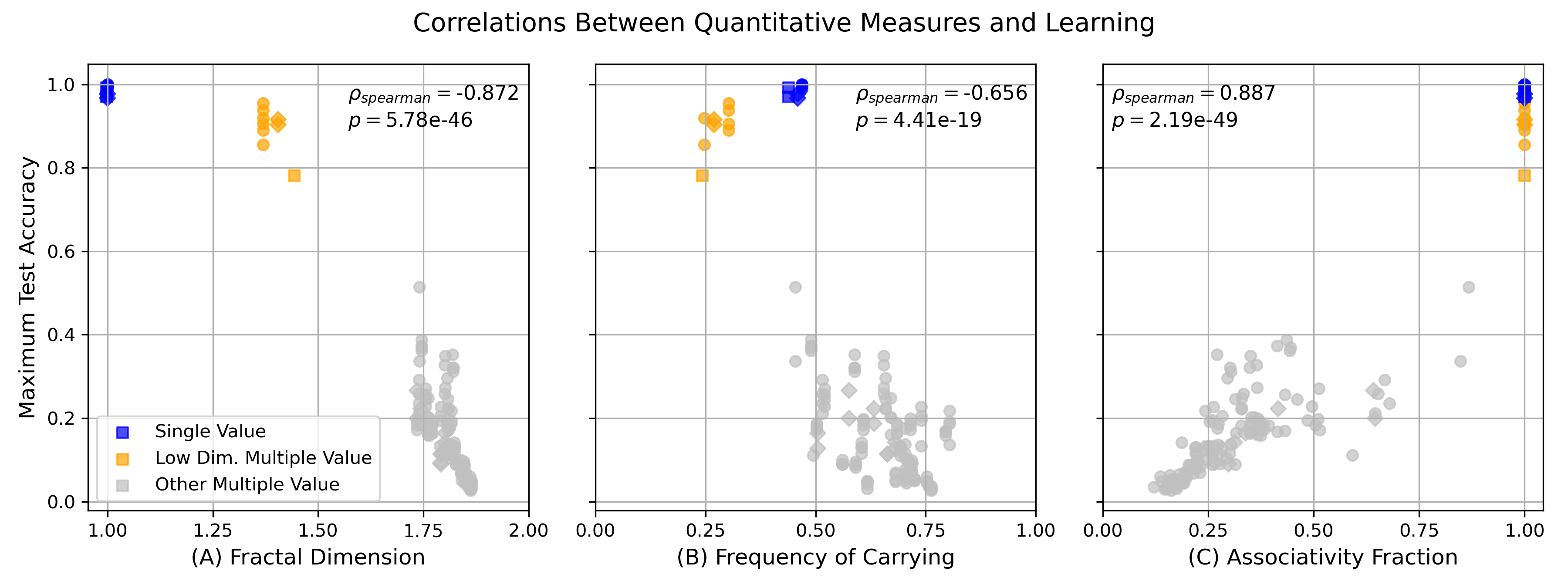}
    \caption{
    Relationship of carry function structure and learning, as measured by maximum testing accuracy (on 6-digit numbers).
    Scatter plots showing relationship of learning curves to structure measured as (A) fractal dimension, (B) frequency of carrying, and (C) associativity fraction for different carry functions, divided into three categories: Single Value carry functions (blue), Low  Dimensional Multiple Value carry functions (orange), and other Multiple Value carry functions (grey). 
    The base $b$ of each carry function is indicated by shape ($b=3$: squares; $b=4$: diamonds; and $b=5$: circles).
    Spearman's rank correlations (over bases $b=3$ to 5) and significance are shown for each plot.
    }
    \label{fig:7}
\end{figure}

To evaluate the influence of carry function structure on learning, we measured the correlation between maximum testing accuracy and the quantitative measures described in Section \ref{sec:3} across the carry functions of bases $b=3$ to 5.
Figure~\ref{fig:7} shows scatter plots of maximum testing accuracy (on 6-digit numbers) and the three quantitative measures: fractal dimension, frequency of carrying, and associativity fraction.
The Spearman rank correlation was strongly significant for each measure ($-0.872$, $-0.656$, and $0.887$ for fractal dimension, frequency of carrying, and associativity function, respectively).
The correlations suggest that fractal dimension and frequency of carrying have a negative influence on learning, and associativity fraction a positive influence.
Furthermore, across all measures, there was a clear separation into the same three groupings of Single Value carry functions, Low Dimensional Multiple Value carry functions, and other Multiple Value carry function.
We discuss these findings further in Section~\ref{sec:5}.

For robustness, we used two additional measurements of learning: the upper asymptote and critical point of a sigmoid function fitted to the testing accuracy.
For more details including correlations between these metrics and the quantitative measures, see Appendix~\hyperref[app:C.2]{C.2}.

\section{Discussion} \label{sec:5}

In this paper, we presented a group theoretic analysis of base addition, that identified two fundamental classes of carry functions: Single Value  and Multiple Value.
We quantified their structure using measures of fractal dimension, frequency of carrying, and associativity fraction and found, across all measures, a clear separation between Single Value and Low Dimensional Multiple Value versus other Multiple Value carry functions.
In neural network simulations, these separated into three groups that aligned with the three types of carry functions with respect to the effectiveness with which they were learned.
Here, we discuss the implications of these results in terms of mathematics and neural networks.

\subsection{Mathematics} \label{sec:5.1}

The analyses reported here extend a nascent line of work considering base arithmetic from a group theoretic perspective.
Specifically, they help formalize and characterize the nature of the symmetries underlying base addition, that can serve as a foundation for further work that extends the analysis in a number of ways.
Most proximally, questions remain about the nature of  Multiple Value carry functions: do these correspond to some factorization of the base representation, and which if any are $\infty$-equivariant?
For Low Dimensional Multiple Value carry functions, all of which are all at least 5-equivariant (see Figure~\ref{fig:3}C), is there a factorization of their base representations that can explain their differences from the other Multiple Value carry functions and, more generally, is the scale of a base addition's equivariance related to some factorization of its associated base representation?
Extending our analyses to the more general case of addition (i.e., over arbitrary numbers of addends, that may involve more complex forms of carry), and to other arithmetic operations are also important directions for future work.

As noted in Section~\ref{sec:1}, characterizing base arithmetic in group theoretic terms provides an opportunity for formalizing our understanding of how human cognition exploits symmetries for generalization.
The work presented here, on base addition, should be directly relevant to a formal understanding of factors such as translational invariance.
More broadly, it would be interesting to explore the extent to which other basic mathematical operations  that are central to human cognitive function -- such as rotations, processing of hierarchical structures (such as trees), and context-normalization -- can build on the work presented here and/or be addressed using a similar approach.

\subsection{Neural Networks} \label{sec:5.2}

\subsubsection{Symmetry Discovery} \label{sec:5.2.1}

Notable among the results in Section~\ref{sec:4} is the capacity for a very small model (1-layer GRU) to learn addition with all of the Single Value carry functions and generalize to considerably more digits than the number on which it was trained.
For all of these carry functions, the neural network achieved near-perfect accuracy on numbers with up to twice as many digits (6) as the training set (3),
and maintained high accuracy on numbers with as many as 10 digits.
This no doubt reflects the underlying symmetry of the carry functions, and likely the ability of a recurrent neural network to discover that given the inductive bias of weight sharing over time, a factor we discuss in the sections that follow.
It is also a direct result of the way in which the problem was constructed and presented to the network, a factor that we discuss further in Section~\ref{sec:5.2.3}.
Collectively, these factors may explain where and how the human brain supports the radical generalization abilities evident in human cognition.

\subsubsection{Structure and Learning} \label{sec:5.2.2}

The results presented in Section~\ref{sec:4} indicated that the effectiveness with which a neural network learns a carry function is directly impacted by several quantifiable features of its structure.
The most obvious and straightforward is the complexity of the function, measured as its fractal dimension.
This is not surprising, as it is to be expected that a neural network finds it more difficult to learn more complex functions \citep{Loukas2021, Goldblum2023}.
This may also explain why the carry function used universally by humans -- the $\mathbf{1}$ carry function -- is a Single Value carry function: they are the simplest to learn.
Furthermore, the simulations reported in Section~\ref{sec:4.2.1} suggest why the $\mathbf{1}$ carry function in particular is so universal: not only is it a Single Value carry function, but it is also aligned with the ordinal structure of numbers.
The latter must be encoded using some form of semantic structure, which, as shown in Figure \ref{fig:5}, makes it comparatively easier to learn than other Single Value carry functions.

We also found that, for the more easily learned carry functions  -- Single Value and Low Dimensional Multiple Value -- there was a positive relationship between frequency of carries and the maximum testing accuracy (evidenced by the blue and orange dots in Figure~\ref{fig:7}B).
This is not surprising, insofar as higher frequency of occurrence provides more opportunities to learn a given operation.
Interestingly, however, this was not so for the more complex Multiple Value carry functions which, despite considerably higher frequencies of carry, exhibited considerably longer learning times.
This is likely due to the fact that, although carries are frequent in those functions, the specific values being carried vary considerably, thus diminishing the benefits of repetition.

Perhaps most importantly, the carry functions with greatest associativity — that is, the most compact forms of symmetry -- are learned the most effectively, as shown in Figure~\ref{fig:7}C.
This could provide a formal account for why humans universally use the $\mathbf{1}$ carry function, and more generally for the importance of symmetries for learning in neural networks.

Finally, we return to the observation, in Figure~\ref{fig:6}, that accuracy improved with base size.  On the surface, this might seem surprising, as carry tables for larger bases might be expected to be more complex and therefore more difficult to learn.  
However, a simple scaling analysis reveals that, while the complexity of modular addition and the carry function are upper-bounded by $b^2$ (i.e., memorization for each pair), the number of training examples scales with $b^6$ (i.e., all pairs of 3-digit numbers).
Hence, the benefit of more training examples may outweigh the greater difficulty of learning the addition function in that base, by leveraging the underlying symmetries for more effective learning -- much as people seem to do for base 10.

\subsubsection{Curriculum} \label{sec:5.2.3}

As discussed in Section~\ref{sec:1}, the way in which a network is trained can be as important an inductive bias as the architecture itself (e.g., recurrence) for discovering symmetries.
Here, we exploited this by presenting the data to the network in a format that made the symmetry as apparent and accessible as possible: one digit at a time, from least to most significant.
This made it possible for the network to learn the modulus arithmetic and carry operations, without having to discover the relevance of and learn to manage digit order.
However, this leaves open the question of how a network may learn to perform base arithmetic when this is presented in a less constrained form, such as the most familiar notation to humans (in which all of the digits of one number are presented before the other).
This would require it to parse the problem into a form that exposes the underlying symmetry, as well as learning to perform the other ancillary operations, such as appropriately ordering the digits for processing, caching intermediate values, and reordering them for the response.
Discovering and coordinating the use of these functions turns out to be a challenging problem for neural networks, with no clear evidence that even the largest current models are able to do so in a way that supports fully algorithmic base arithmetic \citep{Nogueira2021, Zhou2022, Qian2022, Ebrahimi2024}.
This suggests that other architectural elements may be important, and even necessary as inductive biases for the discovery and implementation of such functions, such as access to an external memory and position coding.
Work in cognitive science suggests that humans make use of such functions in problems that involve memory, sequential processing, and numerical comparisons (e.g., \cite{Howard2002, Beukers2024}).
Integrating these into the model we have presented here for discovery of the  symmetries underlying base arithmetic offers a promising direction for future research.
\section{Summary and Conclusions} \label{sec:6}

In this paper, we provide a formal group theoretical analysis of base addition in terms of the symmetry properties of underlying carry functions.
We quantify three features of such carry functions with respect to multi-digit addition: their complexity (measured as fractal dimension), the frequency of carries, and the compactness of the underlying symmetry functions (or their degree of equivariance).
In all cases, we find a close relationship between these properties and the ease with which a neural network can learn the corresponding functions, and discuss how these observations may provide insights into why humans universally use a particular form of carry -- the $\mathbf{1}$ carry function -- in base addition: it is the least complex, with highest frequency of carries, and exhibits the most compactness (highest degree of equivariance).
This is supported by the observation that this is the form of carry more quickly and effectively learned in a simple recurrent neural network.
We also consider the importance of the format in which training is carried out, again providing evidence that neural networks learn most effectively under the same training paradigm that is most commonly used to teach multi-digit addition to humans.

More generally, the work presented here provides an example of how the learning of underlying symmetries undergirds the capacity for radical generalization that is one of the defining characteristics of human cognitive function.
Our approach offers a formal framework for characterizing such symmetries, and studying how they can be learned in neural networks.
In particular, it illustrates how inductive biases -- both structural (such as recurrence) and curricular (here, the interleaved training protocol) can help make symmetries much more accessible for discovery, thus greatly enhancing the sample efficiency of the network.
Such insights promise to be valuable both for understanding how people learn and make use of other forms of symmetry structure, as well as the design of artificial systems that do so more efficiently than existing systems.

\section*{Acknowledgements}

The authors would like to thank Mark McConnell for helping to formalize the mathematical construction of base addition, as well as the following individuals for useful conversations and suggestions regarding the work throughout this manuscript: Taylor Webb, Zack Dulberg, Adel Ardalan, Tim Buschman, and Shanka Mondal.
SS was supported by a T32 Training Grant in Computational Neuroscience (T32MH065214), and KK was supported by a C. V. Starr fellowship from the Princeton Neuroscience Institute and a CPBF (Center for the Physics of Biological Function) fellowship (through NSF PHY-1734030).
This work was supported in part by a Vannevar Bush Faculty Fellowship from the Office of the Under Secretary of Defense for Research \& Engineering (ONR N00014-22-1-2002) awarded to JDC.

\bibliography{references}

\begin{thebibliography}{26}
\expandafter\ifx\csname natexlab\endcsname\relax\def\natexlab#1{#1}\fi
\providecommand{\url}[1]{\texttt{#1}}
\providecommand{\href}[2]{#2}
\providecommand{\path}[1]{#1}
\providecommand{\DOIprefix}{doi:}
\providecommand{\ArXivprefix}{arXiv:}
\providecommand{\URLprefix}{URL: }
\providecommand{\Pubmedprefix}{pmid:}
\providecommand{\doi}[1]{\href{http://dx.doi.org/#1}{\path{#1}}}
\providecommand{\Pubmed}[1]{\href{pmid:#1}{\path{#1}}}
\providecommand{\bibinfo}[2]{#2}
\ifx\xfnm\relax \def\xfnm[#1]{\unskip,\space#1}\fi
\bibitem[{Anderson(1978)}]{Anderson1978}
\bibinfo{author}{Anderson, P.W.}, \bibinfo{year}{1978}.
\newblock \bibinfo{title}{Local moments and localized states}.
\newblock \bibinfo{journal}{Science} \bibinfo{volume}{201}, \bibinfo{pages}{307--316}.
\bibitem[{Beukers et~al.(2024)Beukers, Hamin, Norman and Cohen}]{Beukers2024}
\bibinfo{author}{Beukers, A.}, \bibinfo{author}{Hamin, M.}, \bibinfo{author}{Norman, K.A.}, \bibinfo{author}{Cohen, J.D.}, \bibinfo{year}{2024}.
\newblock \bibinfo{title}{When working memory may be just working, not memory.}
\newblock \bibinfo{journal}{Psychological Review} \bibinfo{volume}{131}, \bibinfo{pages}{563}.
\bibitem[{Brown(2012)}]{Brown2012}
\bibinfo{author}{Brown, K.}, \bibinfo{year}{2012}.
\newblock \bibinfo{title}{Cohomology of Groups}.
\newblock Graduate Texts in Mathematics, \bibinfo{publisher}{Springer New York}.
\bibitem[{Cho et~al.(2014)Cho, Van~Merri{\"e}nboer, Gulcehre, Bahdanau, Bougares, Schwenk and Bengio}]{Cho2014}
\bibinfo{author}{Cho, K.}, \bibinfo{author}{Van~Merri{\"e}nboer, B.}, \bibinfo{author}{Gulcehre, C.}, \bibinfo{author}{Bahdanau, D.}, \bibinfo{author}{Bougares, F.}, \bibinfo{author}{Schwenk, H.}, \bibinfo{author}{Bengio, Y.}, \bibinfo{year}{2014}.
\newblock \bibinfo{title}{Learning phrase representations using rnn encoder-decoder for statistical machine translation}.
\newblock \bibinfo{journal}{arXiv preprint arXiv:1406.1078} .
\bibitem[{Cleeremans and McClelland(1991)}]{cleeremans1991learning}
\bibinfo{author}{Cleeremans, A.}, \bibinfo{author}{McClelland, J.L.}, \bibinfo{year}{1991}.
\newblock \bibinfo{title}{Learning the structure of event sequences.}
\newblock \bibinfo{journal}{Journal of Experimental Psychology: General} \bibinfo{volume}{120}, \bibinfo{pages}{235}.
\bibitem[{Dehaene(1997)}]{Dehaene1997}
\bibinfo{author}{Dehaene, S.}, \bibinfo{year}{1997}.
\newblock \bibinfo{title}{The Number Sense: How the Mind Creates Mathematics}.
\newblock \bibinfo{publisher}{Oxford University Press}, \bibinfo{address}{New York}.
\bibitem[{Ebrahimi et~al.(2024)Ebrahimi, Panchal and Memisevic}]{Ebrahimi2024}
\bibinfo{author}{Ebrahimi, M.}, \bibinfo{author}{Panchal, S.}, \bibinfo{author}{Memisevic, R.}, \bibinfo{year}{2024}.
\newblock \bibinfo{title}{Your context is not an array: Unveiling random access limitations in transformers}.
\newblock \bibinfo{journal}{arXiv preprint arXiv:2408.05506} .
\bibitem[{Elman(1990)}]{Elman1990}
\bibinfo{author}{Elman, J.L.}, \bibinfo{year}{1990}.
\newblock \bibinfo{title}{Finding structure in time}.
\newblock \bibinfo{journal}{Cognitive Science} \bibinfo{volume}{14}, \bibinfo{pages}{179--211}.
\newblock \DOIprefix\doi{https://doi.org/10.1207/s15516709cog1402\_1}.
\bibitem[{Falconer(2014)}]{Falconer2014}
\bibinfo{author}{Falconer, K.}, \bibinfo{year}{2014}.
\newblock \bibinfo{title}{Fractal Geometry: Mathematical Foundations and Applications}.
\newblock \bibinfo{publisher}{John Wiley \& Sons}.
\bibitem[{Goldblum et~al.(2023)Goldblum, Finzi, Rowan and Wilson}]{Goldblum2023}
\bibinfo{author}{Goldblum, M.}, \bibinfo{author}{Finzi, M.}, \bibinfo{author}{Rowan, K.}, \bibinfo{author}{Wilson, A.G.}, \bibinfo{year}{2023}.
\newblock \bibinfo{title}{The no free lunch theorem, kolmogorov complexity, and the role of inductive biases in machine learning}.
\newblock \bibinfo{journal}{arXiv preprint arXiv:2304.05366} .
\bibitem[{Hochreiter and Schmidhuber(1997)}]{Hochreiter1997}
\bibinfo{author}{Hochreiter, S.}, \bibinfo{author}{Schmidhuber, J.}, \bibinfo{year}{1997}.
\newblock \bibinfo{title}{{Long Short-Term Memory}}.
\newblock \bibinfo{journal}{Neural Computation} \bibinfo{volume}{9}, \bibinfo{pages}{1735--1780}.
\newblock \DOIprefix\doi{10.1162/neco.1997.9.8.1735}, \href{http://arxiv.org/abs/https://direct.mit.edu/neco/article-pdf/9/8/1735/813796/neco.1997.9.8.1735.pdf}{{\tt arXiv:https://direct.mit.edu/neco/article-pdf/9/8/1735/813796/neco.1997.9.8.1735.pdf}}.
\bibitem[{Howard and Kahana(2002)}]{Howard2002}
\bibinfo{author}{Howard, M.W.}, \bibinfo{author}{Kahana, M.J.}, \bibinfo{year}{2002}.
\newblock \bibinfo{title}{A distributed representation of temporal context}.
\newblock \bibinfo{journal}{Journal of mathematical psychology} \bibinfo{volume}{46}, \bibinfo{pages}{269--299}.
\bibitem[{Isaksen(2002)}]{Isaksen2002}
\bibinfo{author}{Isaksen, D.}, \bibinfo{year}{2002}.
\newblock \bibinfo{title}{A cohomological viewpoint on elementary school arithmetic}.
\newblock \bibinfo{journal}{The American Mathematical Monthly} .
\bibitem[{Ji-An et~al.(2024)Ji-An, Benna and Mattar}]{Ji-An2023}
\bibinfo{author}{Ji-An, L.}, \bibinfo{author}{Benna, M.K.}, \bibinfo{author}{Mattar, M.G.}, \bibinfo{year}{2024}.
\newblock \bibinfo{title}{Discovering cognitive strategies with tiny recurrent neural networks}.
\newblock \bibinfo{journal}{bioRxiv} \URLprefix \url{https://www.biorxiv.org/content/early/2024/10/05/2023.04.12.536629}, \DOIprefix\doi{10.1101/2023.04.12.536629}, \href{http://arxiv.org/abs/https://www.biorxiv.org/content/early/2024/10/05/2023.04.12.536629.full.pdf}{{\tt arXiv:https://www.biorxiv.org/content/early/2024/10/05/2023.04.12.536629.full.pdf}}.
\bibitem[{Krizhevsky et~al.(2012)Krizhevsky, Sutskever and Hinton}]{Krizhevsky2012}
\bibinfo{author}{Krizhevsky, A.}, \bibinfo{author}{Sutskever, I.}, \bibinfo{author}{Hinton, G.E.}, \bibinfo{year}{2012}.
\newblock \bibinfo{title}{Imagenet classification with deep convolutional neural networks}, in: \bibinfo{booktitle}{Advances in Neural Information Processing Systems}, \bibinfo{publisher}{Curran Associates, Inc.}
\bibitem[{Lecun et~al.(1998)Lecun, Bottou, Bengio and Haffner}]{LeCun1998}
\bibinfo{author}{Lecun, Y.}, \bibinfo{author}{Bottou, L.}, \bibinfo{author}{Bengio, Y.}, \bibinfo{author}{Haffner, P.}, \bibinfo{year}{1998}.
\newblock \bibinfo{title}{Gradient-based learning applied to document recognition}.
\newblock \bibinfo{journal}{Proceedings of the IEEE} \bibinfo{volume}{86}, \bibinfo{pages}{2278--2324}.
\newblock \DOIprefix\doi{10.1109/5.726791}.
\bibitem[{Loukas et~al.(2021)Loukas, Poiitis and Jegelka}]{Loukas2021}
\bibinfo{author}{Loukas, A.}, \bibinfo{author}{Poiitis, M.}, \bibinfo{author}{Jegelka, S.}, \bibinfo{year}{2021}.
\newblock \bibinfo{title}{What training reveals about neural network complexity}.
\newblock \bibinfo{journal}{Advances in Neural Information Processing Systems} \bibinfo{volume}{34}, \bibinfo{pages}{494--508}.
\bibitem[{Nogueira et~al.(2021)Nogueira, Jiang and Lin}]{Nogueira2021}
\bibinfo{author}{Nogueira, R.}, \bibinfo{author}{Jiang, Z.}, \bibinfo{author}{Lin, J.}, \bibinfo{year}{2021}.
\newblock \bibinfo{title}{Investigating the limitations of transformers with simple arithmetic tasks}.
\newblock \bibinfo{journal}{arXiv preprint arXiv:2102.13019} .
\bibitem[{Piantadosi(2023)}]{Piantadosi2023}
\bibinfo{author}{Piantadosi, S.T.}, \bibinfo{year}{2023}.
\newblock \bibinfo{title}{The algorithmic origins of counting}.
\newblock \bibinfo{journal}{Child Development} \bibinfo{volume}{94}, \bibinfo{pages}{1472--1490}.
\bibitem[{Qian et~al.(2022)Qian, Wang, Li, Li and Yan}]{Qian2022}
\bibinfo{author}{Qian, J.}, \bibinfo{author}{Wang, H.}, \bibinfo{author}{Li, Z.}, \bibinfo{author}{Li, S.}, \bibinfo{author}{Yan, X.}, \bibinfo{year}{2022}.
\newblock \bibinfo{title}{Limitations of language models in arithmetic and symbolic induction}.
\newblock \bibinfo{journal}{arXiv preprint arXiv:2208.05051} .
\bibitem[{Segert(2024)}]{Segert2024}
\bibinfo{author}{Segert, S.}, \bibinfo{year}{2024}.
\newblock \bibinfo{title}{Maximum Entropy, Symmetry, and Relational Bottleneck: Unraveling the Impact of Inductive Biases on Systematic Reasoning}.
\newblock Ph.D. thesis. Princeton University.
\bibitem[{Segert and Cohen(2022)}]{segert2022a}
\bibinfo{author}{Segert, S.}, \bibinfo{author}{Cohen, J.}, \bibinfo{year}{2022}.
\newblock \bibinfo{title}{A self-supervised framework for function learning and extrapolation}.
\newblock \bibinfo{journal}{Transactions on Machine Learning Research} \URLprefix \url{https://openreview.net/forum?id=ILPFasEaHA}.
\bibitem[{Webb et~al.(2023)Webb, Holyoak and Lu}]{Webb2023}
\bibinfo{author}{Webb, T.}, \bibinfo{author}{Holyoak, K.J.}, \bibinfo{author}{Lu, H.}, \bibinfo{year}{2023}.
\newblock \bibinfo{title}{Emergent analogical reasoning in large language models}.
\newblock \bibinfo{journal}{Nature Human Behavior} .
\bibitem[{Wynn(1992)}]{Wynn1992}
\bibinfo{author}{Wynn, K.}, \bibinfo{year}{1992}.
\newblock \bibinfo{title}{Addition and subtraction by human infants}.
\newblock \bibinfo{journal}{Nature} \bibinfo{volume}{358}, \bibinfo{pages}{749--750}.
\bibitem[{Yang et~al.(2025)Yang, Campbell, Huang, Wang, Cohen and Webb}]{Yang2025}
\bibinfo{author}{Yang, Y.}, \bibinfo{author}{Campbell, D.}, \bibinfo{author}{Huang, K.}, \bibinfo{author}{Wang, M.}, \bibinfo{author}{Cohen, J.}, \bibinfo{author}{Webb, T.}, \bibinfo{year}{2025}.
\newblock \bibinfo{title}{Emergent symbolic mechanisms support abstract reasoning in large language models}.
\newblock \URLprefix \url{https://arxiv.org/abs/2502.20332}, \href{http://arxiv.org/abs/2502.20332}{{\tt arXiv:2502.20332}}.
\bibitem[{Zhou et~al.(2022)Zhou, Nova, Larochelle, Courville, Neyshabur and Sedghi}]{Zhou2022}
\bibinfo{author}{Zhou, H.}, \bibinfo{author}{Nova, A.}, \bibinfo{author}{Larochelle, H.}, \bibinfo{author}{Courville, A.}, \bibinfo{author}{Neyshabur, B.}, \bibinfo{author}{Sedghi, H.}, \bibinfo{year}{2022}.
\newblock \bibinfo{title}{Teaching algorithmic reasoning via in-context learning}.
\newblock \bibinfo{journal}{arXiv preprint arXiv:2211.09066} .

\end{thebibliography}
\bibliographystyle{formatting/elsarticle-harv}

\appendix
\newpage

\section{Supplement to Section~\ref{sec:2}: Mathematics of Symmetry and Base Addition} \label{app:A}

\subsection{Supplemental Group Theory} \label{app:A.1}

Two foundational concepts of group theory relevant to this work are \textit{homomorphism} and \textit{isomorphism}: Given two groups $G$ and $G'$, a group homomorphism is a mapping $f: G \to G'$ that preserves the structure of both groups (i.e., $f(g \cdot h) = f(g) \cdot f(h)$ for all $g, h \in G$); and a group isomorphism is a homomorphism that is also bijective (i.e., forms a one-to-one mapping).
Homomorphisms and isomorphisms formalize what it means for two groups to have similar or the same structure, respectively, and hence allow for comparing and classifying groups.
For example, let $G'$ be the group of integers modulo 3, with elements $0, 1, 2$ and a group operation of addition modulo 3 (e.g., $2 + 2 = 1$).
In Section~\ref{sec:2.1}, we mention that $G'$ is isomorphic to the rotational group of the equilateral triangle (which we denote by $G$).
To see this, let $f: G \to G'$ map the rotations of $0\degree, 120\degree, 240\degree$ clockwise to $0, 1, 2$, respectively.
This reflects that, though seemingly entirely different on the surface, the two groups are different expressions of the same underlying symmetry.

Another important construct in group theory is that of a \textit{group action}, in which $G$ ``acts'' on a set $X$ in a way that respects the structure of $G$.
For example, the rotational group of the triangle may act on the Euclidean plane by rotation about the origin.
The \textit{orbit} of an element $x \in X$ denotes the subset of $X$ that $x$ is sent to by all $g \in G$; e.g., the orbit of $(1, 0)$ in the preceding example is $\{ (1, 0), (-\frac{1}{2}, \frac{\sqrt{3}}{2}), (-\frac{1}{2}, -\frac{\sqrt{3}}{2}) \}$.
Similarly, the orbit $G \cdot A$ of a subset $A \subset X$ denotes the subset of $X$ to which all $x \in A$ are sent by all $g \in G$.
A group action serves to separate a space's symmetry from the space itself, and provides a framework with which to compare different spaces with the same symmetry.

\subsection{Formal Derivation of Base Addition and Carry Functions} \label{app:A.2}

In this appendix, we derive the formalism underpinning base addition.

Choose a base $b \in \N_{\geq 2}$, and denote the integers modulo $b$ by $\Z_b$.
For any integer $n \in \Z$, we wish to present it in a \textit{base representation}; i.e., we seek a tuple $(n_k, n_{k-1}, ..., n_1)$, where $n_j \in \Z_b$ is the $j\textsuperscript{th}$ digit for $j \in [k]$.
To properly preserve the structure of $\Z$, we will construct base representations with use of concepts from group cohomology, following \citet{Isaksen2002}.
Note that the material and exposition in Appendix~\hyperref[app:A.2.1]{A.2.1}, Appendix~\hyperref[app:A.2.2]{A.2.2}, and Appendix~\hyperref[app:A.2.3]{A.2.3} closely follows \citet{Brown2012}, which establishes the formalism for the 2-digit case.
In Appendix~\hyperref[app:A.2.4]{A.2.4}, we iteratively extend to arbitrary digits.

\textit{Note}:
Throughout, in a slight abuse of notation (and because we only consider abelian groups), we use ``$+$'' to jointly refer to group operations of different groups, ``$-n$'' to refer to inverse of $n$, and ``$m \cdot n$'' to refer to the multiple $n = \underbrace{n + \cdots + n}_{m \text{ times}}$.

\subsubsection{Group Extensions} \label{app:A.2.1}

To start, we seek to represent $n \in \Z_{b^2}$ as a 2-digit number $(n_2, n_1) \in \Z_b \times \Z_b$.
We may approach this as a group extension problem, framing $\Z_{b^2}$ as an extension of $\Z_b$ by itself.
Thus, we have the short exact sequence
\begin{equation} \label{eq:short-exact-sequence}
    0 \to \Z_b \xrightarrow{i} \Z_{b^2} \xrightarrow{\pi} \Z_b \to 0,
\end{equation}
where $i: \Z_b \to \Z_{b^2}$ is the inclusion map of the second digit and $\pi: \Z_{b^2} \to \Z_b$ is the projection map to the first digit.

To properly construct a base representation, we seek an extension equivalent to $\Z_{b^2}$ defined on the set $\Z_b \times \Z_b$; i.e., an appropriate group law on $\Z_b \times \Z_b$ (we denote the resulting group by $\Z_b \times_f \Z_b$), new inclusion and projection maps $\hat{\imath}$ and $\hat{\pi}$, and an isomorphism $\phi_s: \Z_b \times_f \Z_b \to \Z_{b^2}$ making the diagram in Figure~\ref{fig:A8} commute (we explain these notations in Appendix~\hyperref[app:A.2.2]{A.2.2}).

\begin{figure}[h] 
    \centering
    \begin{tikzcd}[row sep=tiny]
        & & \Z_{b^2} \arrow[rd, "\pi"] \\
        0 \arrow[r] & \Z_b \arrow[ur, "i"] \arrow[dr, "\hat{\imath}"] & & \Z_b \arrow[r] & 0 \\
        & & \Z_b \times_f \Z_b \arrow[uu, "\phi_s"] \arrow[ru, "\hat{\pi}"]
    \end{tikzcd}
    \caption{The base $b$ representation $\Z_b \times_f \Z_b$ is equivalent to $\Z_{b^2}$ with proper choice of section $s$ and cocycle $f$ (see Appendix~\hyperref[app:A.2.2]{A.2.2}).}
    \label{fig:A8}
\end{figure}

\subsubsection{Constructing an Equivalent Extension} \label{app:A.2.2}

We seek to construct a group law on $\Z_b \times \Z_b$ such that it is equivalent to $\Z_{b^2}$.
We will see that this reduces to choosing a way to ``carry'' from the first  to second digits when adding, which we will define in purely group cohomological terms.

First, choose a set-theoretic cross-section of the projection $\pi$, i.e., a function $s: \Z_b \to \Z_{b^2}$ such that $\pi s = \textnormal{id}_{\Z_b}$.
Assume that $s$ satisfies the \textit{normalization condition}, $s(0) = 0$.
Note that $s$ may or may not be a homomorphism; to measure the failure of $s$ to be one, we define $f: \Z_b \times \Z_b \to \Z_b$ by
\begin{equation*}
    s(n) + s(m) = i(f(n, m)) + s(n + m).
\end{equation*}
Notice $f$ is well-defined because, for any $n, m \in \Z_b$, $s(n) + s(m)$ and $s(n + m)$ differ exactly by some element in $i(\Z_b)$ and $i$ is injective.
This definition also implies that $f$ is \textit{normalized}; i.e., $f(n, 0) = 0 = f(0, n)$ for all $n \in \Z_b$.
In the spirit of base addition, we will sometimes refer to $f$ as the \textit{carry function} associated with $s$.

We may use any such $s$ to: (i) construct a bijection $\phi_s$ between $\Z_b \times \Z_b$ and $\Z_{b^2}$; and (ii) compute the group law on $\Z_b \times \Z_b$ such that $\phi_s$ is an isomorphism.
Notice that $s(\Z_b)$ is a set of coset representatives for $i(\Z_b)$ in $\Z_{b^2}$.
Therefore, $\phi_s: \Z_b \times \Z_b \to \Z_{b^2}$ defined by $\phi_s(n_2, n_1) = i(n_2) + s(n_1)$ is a bijection.
Next, for two elements $(n_2, n_1), (m_2, m_1) \in \Z_b \times \Z_b$, we may write
\begin{align*}
    \phi_s(n_2, n_1) + \phi_s(m_2, m_1)
        &= i(n_2) + s(n_1) + i(m_2) + s(m_1) \\
        &= i(n_2) + i(m_2) + s(n_1) + s(m_1) \\
        &= i(n_2) + i(m_2) + i(f(n_1, m_1)) + s(n_1 + m_1) \\
        &= i(n_2 + m_2 + f(n_1, m_1)) + s(n_1 + m_1),
\end{align*}
where in the second equality we have used that $\Z_{b^2}$ is abelian.
Choosing the group law on $\Z_b \times \Z_b$
\begin{equation} \label{eq:group-law}
    (n_2, n_1) + (m_2, m_1) = (n_2 + m_2 + f(n_1, m_1), n_1 + m_1),
\end{equation}
then $\phi_s$ is an isomorphism.
We henceforth use $\Z_b \times_f \Z_b$ to denote $\Z_b \times \Z_b$ with the group law (\ref{eq:group-law}) and carry function $f$.

For our inclusion and projection maps, define $\hat{\imath}: \Z_b \to \Z_b \times \Z_b$ by $\hat{\imath}(n) = (n, 0)$, and $\hat{\pi}: \Z_b \times \Z_b \to \Z_b$ by $\hat{\pi}(n_2, n_1) = n_1$.
Notice $\hat{\imath}$ is just the canonical inclusion map of the second digit, and $\hat{\pi}$ the canonical projection map to the first digit.
With $\hat{\imath}$, $\hat{\pi}$ as defined, Figure~\ref{fig:A8} commutes for any choice of section $s$.
For any $n \in \Z_b$, $(n_2, n_1) \in \Z_b \times_f \Z_b$,
\begin{gather*}
    \phi_s(\hat{\imath}(n)) = \phi_s((n, 0)) = i(n), \\
    \pi(\phi_s((n_2, n_1))) = \pi(i(n_2) + s(n_1)) = n_1 = \pi(n_1).
\end{gather*}
Therefore, $\Z_b \times_f \Z_b$ is equivalent to $\Z_{b^2}$, and we have established how to properly represent numbers in $\{0, 1, ..., b^2 - 1\}$ as two digits, each in $\{0, 1, ..., b\}$.

Note that not any function $f: \Z_b \times \Z_b \to \Z_b$ defines a group $\Z_b \times_f \Z_b$.
In fact, $\Z_b \times_f \Z_b$ is associative if and only if $f$ satisfies
\begin{equation} \label{eq:cocycle-condition}
    f(n, m) + f(n + m, p) = f(m, p) + f(n, m + p).
\end{equation}
For, choosing arbitrary $(n_2, n_1), (m_2, m_1), (p_2, p_1) \in \Z_b \times_f \Z_b$, then
\begin{align*}
    \big((n_2, n_1) + (m_2,& m_1)\big) + (p_2, p_1) \\
        &= (n_2 + m_2 + f(n_1, m_1), n_1 + m_1) + (p_2, p_1) \\
        &= (n_2 + m_2 + p_2 + f(n_1, m_1) + f(n_1 + m_1, p_1), n_1 + m_1 + p_1), \\
    (n_2, n_1) + \big((m_2,& m_1) + (p_2, p_1)\big) \\
        &= (n_2, n_1) + (m_2 + p_2 + f(m_1, p_1), m_1 + p_1) \\
        &= (n_2 + m+2 + p_2 + f(m_1, p_1) + f(n_1, m_1 + p_1), n_1 + m_1 + p_1),
\end{align*}
which are equal in the second digit exactly when $f$ satisfies (\ref{eq:cocycle-condition}).
We call (\ref{eq:cocycle-condition}) the \textit{cocycle condition} because it implies that $f$ is a 2-cocycle of $\Z_b$ with coefficients in $\Z_b$.

\subsubsection{Comparing Equivalent Extensions} \label{app:A.2.3}

For any normalized section $s$, we have shown that $\Z_b \times_f \Z_b$ is equivalent to $\Z_{b^2}$ defined entirely by the associated cocycle $f$ (as well as canonical inclusion and projection maps $\hat{\imath}$ and $\hat{\pi}$).
Now, we will show that changing our choice of section $s'$ corresponds 1-1 to modifying the cocycle $f$ by a coboundary $\delta c$.
Furthermore, we will show that two extensions $\Z_b \times_f \Z_b$ and $\Z_b \times_{f'} \Z_b$ are equivalent if and only if $f' = f + \delta c$ for some coboundary $\delta c$ (see Figure~\ref{fig:A9} for the commutative diagram).

\begin{figure}[h]
    \centering
    \begin{tikzcd}
        & & \Z_{b^2} \arrow[rd, "\pi"] \\
        0 \arrow[r] & \Z_b \arrow[ur, "i"] \arrow[r, "\hat{\imath}"] \arrow[dr, "\hat{\imath}"] & \Z_b \times_f \Z_b \arrow[u, "\phi_s"] \arrow[d, "\psi_c"] \arrow[r, "\hat{\pi}"] & \Z_b \arrow[r] & 0 \\
        & & \Z_b \times_{f'} \Z_b \arrow[ru, "\hat{\pi}"]
    \end{tikzcd}
    \caption{We have already shown that $\Z_{b^2}$ and $\Z_b \times_f \Z_b$ are equivalent (see Appendix~\hyperref[app:A.2.2]{A.2.2} and Figure~\ref{fig:A8}).
    With $f' = f + \delta c$, $\Z_b \times_{f'} \Z_b$ is also equivalent (see Appendix~\hyperref[app:A.2.3]{A.2.3}).}
    \label{fig:A9}
\end{figure}

Choose an arbitrary normalized section $s': \Z_b \to \Z_{b^2}$.
We may express $s'$ in terms of our original $s$; i.e., $s'(n) = i(c(n)) + s(n)$, where $c:\Z_b \to \Z_b$ satisfies $c(0) = 0$.
To compute the cocycle $f'$ associated with $s'$, we have
\begin{align*}
    s'(n) + s'(m)
        =& \, i(c(n)) + s(n) + i(c(m)) + s(m) \\
        =& \, i(c(n)) + i(c(m)) + s(n) + s(m) \\
        =& \, i(c(n) + c(m)) + i(f(n, m)) + s(n + m) \\
        =& \, i(c(n) + c(m) + f(n, m)) + s(n + m) \\
        =& \, i(c(n) + c(m) + f(n, m) - c(n + m)) \\
         & \, + i(c(n + m)) + s(n + m) \\
        =& \, i(c(n) + c(m) + f(n, m) - c(n + m)) + s'(n + m),
\end{align*}
where the second equality holds because $\Z_{b^2}$ is abelian, and the third by definition of $f$.
Then $\delta c: \Z_b \times \Z_b \to \Z_b$ defined by $\delta c(n, m) = c(n) + c(m) - c(n + m)$ is a 2-coboundary of $\Z_b$ with coefficients in $\Z_b$.
So, the cocycle $f': \Z_b \times \Z_b \to \Z_b$ associated with $s'$ is given by $f' = f + \delta c$.

It remains to show that $\Z_b \times_{f'} \Z_b$ is equivalent to $\Z_b \times_f \Z_b$.
Let $\psi_c: \Z_b \times_f \Z_b \to \Z_b \times_{f'} \Z_b$ be defined by $\psi_c(n_2, n_1) = (n_2 + c(n_1), n_1)$.
We will show that $\psi_c$ is an isomorphism.
First, $\psi_c(0, 0) = (0 + c(0), 0) = (0, 0)$, so $\psi_c$ preserves the identity element.
Moreover, we have
\begin{align*}
    \psi_c((n_2, n_1) + (m_2, m_1))
        &= \psi_c((n_2 + m_2 + f(n_1, m_1), n_1 + m_1)) \\
        &= (n_2 + m_2 + f(n_1, m_1) + c(n_1 + m_1), n_1 + m_1) \\
        &= (n_2 + m_2 + c(n_2) + c(m_1) + f'(n_1, m_1), n_1 + m_1) \\
        &= (n_2 + c(n_1), n_1) + (m_2 + c(m_1), m_1) \\
        &= \psi_c(n_2, n_1) + \psi_c(m_2, m_1),
\end{align*}
where the third equality holds because $f' = f + \delta c$.
So, $\psi_c$ also preserves addition.
Furthermore, Figure~\ref{fig:A9} commutes properly.
For any $n \in \Z_b$, $(n_2, n_1) \in \Z_b \times_f \Z_b$,
\begin{gather*}
    \psi_c(\hat{\imath}(n)) = \psi_c((n, 0)) = (n, 0) = \hat{\imath}(n), \\
    \hat{\pi}(\psi_c((n_2, n_1))) = \hat{\pi}((n_2 + c(n_1), n_1)) = n_1 = \hat{\pi}((n_2, n_1)).
\end{gather*}
Therefore, $\Z_b \times_f \Z_b$ and $\Z_b \times_{f'} \Z_b$ are equivalent.

\subsubsection{Iteratively Extending} \label{app:A.2.4}

Thus far, we've shown how to represent 2-digit numbers with an addition (namely, a way to carry from the first to second digit) that preserves the structure of $\Z_{b^2}$.
For a complete base representation of the integers, we wish to present any number in $\Z$ as an arbitrary-length sequence of digits, and when adding, for the carry function between digits to be somehow consistent or recursive.
With this aim, we proceed by iteratively extending our base representation and, each time, choosing the same cocycle (up to embedding).
We will see that -- if the cocycle meets certain constraints -- this procedure produces a base representation with the exact form we want.
We first extend to three digits and then inductively extend for an arbitrary number of digits.
 
For the 3-digit case, we extend $\Z_{b^2}$ by $\Z_b$ to get $\Z_{b^3}$, similarly to $\Z_{b^2}$ in Appendix~\hyperref[app:A.2.1]{A.2.1}.
Denote the inclusion and projection maps by $i_2: \Z_{b^2} \to \Z_{b^3}$ and $\pi_2: \Z_{b^3} \to \Z_b$, respectively.
As before, we seek an equivalent extension defined on the product $(\Z_b)^3 = \Z_b \times \Z_b \times \Z_b$ so it is a proper 3-digit base representation preserving the structure of $\Z_{b^3}$.
Having already shown that the diagram in Figure~\ref{fig:A8} commutes, we now extend the diagram in Figure~\ref{fig:A10} below.
The remainder of this section is devoted to defining a group law on ($\Z_b)^3$ (we denote the resulting group by $(\Z_b)_f^3$), inclusion and projection maps $\hat{\imath}_2$ and $\hat{\pi}_2$, and an isomorphism $\phi_{s_2}: (\Z_b)_f^3 \to \Z_{b^3}$.

\begin{figure}[h]
    \centering
    \begin{tikzcd}[row sep=tiny]
        & & & \Z_{b^3} \arrow[rdd, "\pi_2"] \\
        & & \Z_{b^2} \arrow[ru, "i_2"] \arrow[rrd, dashed, "\pi"{description, left, xshift=-2.2em}] \\
        0 \arrow[r] & \Z_b \arrow[ur, "i"] \arrow[dr, "\hat{\imath}"] & & & \Z_b \arrow[r] & 0 \\
        & & \Z_b \times_f \Z_b \arrow[uu, "\phi_s"] \arrow[rd, "\hat{\imath}_2"] \arrow[rru, dashed, "\hat{\pi}"{description, left, xshift=-2.7em}] \\
        & & & (\Z_b)_f^3 \arrow[uuuu, crossing over, "\phi_{s_2}"] \arrow[ruu, "\hat{\pi}_2"]
    \end{tikzcd}
    \caption{The base $b$ representation $(\Z_b)_f^3$ is equivalent to $\Z_{b^3}$ with proper choice of 2-equivariant cocycle $f$ (see Appendix~\hyperref[app:A.2.4]{A.2.4}).
    With $\pi$ and $\hat{\pi}$ included as dashed lines, notice that the diagram from Figure~\ref{fig:A8} is embedded as a sub-diagram with the same mappings as before.}
    \label{fig:A10}
\end{figure}
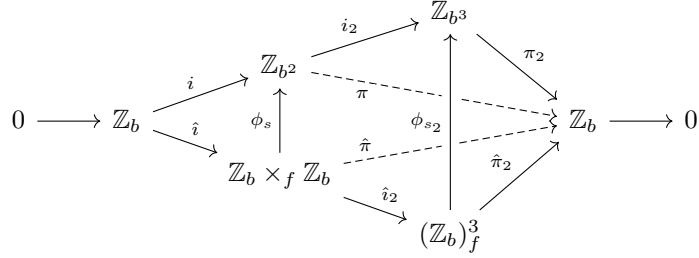

We construct our equivalent representation $(\Z_b)_f^3$ as follows (and recall our section $s: \Z_b \to \Z_{b^2}$, cocycle $f: \Z_b \times \Z_b \to \Z_b$, and $\phi_s: \Z_b \times_f \Z_b \to \Z_{b^2}$ from Appendix~\hyperref[app:A.2.2]{A.2.2}): first, we define section $s_2: \Z_b \to \Z_b^3$, cocycle $f_2: \Z_b \times \Z_b \to \Z_b \times_f \Z_b$, and inclusion and projection maps $\hat{\imath}_2$, $\hat{\pi}_2$ similarly to the 2-digit case; second, we limit our attention to cocycles $f_2$ of the form $(0, f)$; and, third, we rewrite our representation $(\Z_b \times_f \Z_b) \times_{f_2} \Z_b$ as $(\Z_b)_f^3$ with an appropriate group law on the set $(\Z_b)^3$.
Choose a normalized section $s_2: \Z_b \to \Z_{b^3}$ such that $\pi_2 s_2 = \text{id}_{\Z_b}$.
Similarly to before, we define $f_2: \Z_b \times \Z_b \to \Z_b \times_f \Z_b$ by
\begin{equation*}
    s_2(n) + s_2(m) = i_2(\phi_s(f_2(n, m))) + s_2(n + m).
\end{equation*}
$f_2$ is normalized and well-defined by definition of $s_2$, $i_2$, and $\phi_s$ (see Appendix~\hyperref[app:A.2.2]{A.2.2} for details).
Furthermore, if we define the map $\phi_{s_2}: (\Z_b \times_f \Z_b) \times_{f_2} \Z_b \to \Z_{b^3}$ as $\phi_{s_2}((n_3, n_2), n_1) = i_2(\phi_s(n_3, n_2)) + s_2(n_1)$, we have that $\phi_{s_2}$ is an isomorphism if $(\Z_b \times_f \Z_b) \times_{f_2} \Z_b$ has group law
\begin{equation} \label{eq:group-law-ext}
    ((n_3, n_2), n_1) + ((m_3, m_2), m_1)
        = ((n_3, n_2) + (m_3, m_2) + f_2(n_1, m_1), n_1 + m_1),
\end{equation}
which is very similar to (\ref{eq:group-law}).
For our inclusion and projection maps, define $\hat{\imath}_2: \Z_b \times_f \Z_b \to (\Z_b \times_f \Z_b) \times_{f_2} \Z_b$ by $\hat{\imath}_2(n_2, n_1) = ((n_2, n_1), 0)$, and $\hat{\pi}_2: (\Z_b \times_f \Z_b) \times_{f_2} \Z_b \to \Z_b$ by $\hat{\pi}_2((n_3, n_2), n_1) = n_1$.
As with $\hat{\imath}$ and $\hat{\pi}$, notice $\hat{\imath}_2$ is just the canonical inclusion map of the second and third digits, and the $\hat{\pi}_2$ the canonical projection map to the first digit, and it is easy to show that Figure~\ref{fig:A10} commutes for any choice of section $s_2$.
Thus, $(\Z_b \times_f \Z_b) \times_{f_2} \Z_b$ is equivalent to $\Z_{b^3}$.

For the purposes of a base representation, we would like to somehow apply $f$ iteratively rather than having two distinct carries $f$ and $f_2$.
So, suppose $f_2$ is of the form $f_2(n, m) = (0, f(n, m))$.
If $f_2 = (0, f)$ is a cocycle (which is not necessarily the case for arbitrary cocycle $f$), then we call $f$ a \textit{2-equivariant cocycle}.
For any 2-equivariant cocycle $f$, our group law (\ref{eq:group-law-ext}) may be re-written as
\begin{align*}
    \big((n_3,& n_2), n_1\big) + \big((m_3, m_2), m_1\big) \\
        &= \big((n_3, n_2) + (m_3, m_2) + (0, f(n_1, m_1)), n_1 + m_1\big) \\
        &= \big((n_3 + m_3 + f(n_2, m_2), n_2 + m_2) + (0, f(n_1, m_1)), n_1 + m_1\big) \\
        &= \big(n_3 + m_3 + f(n_2, m_2) + f(n_2 + m_2, f(n_1, m_1)), n_2 + m_2 + f(n_1, m_1)), n_1 + m_1\big),
\end{align*}
and we have essentially just defined an addition on the set $(\Z_b)^3$ making it isomorphic to $\Z_{b^3}$.
To be explicit, let our group rule on $(\Z_b)^3$ be
\begin{align*}
    (n_3,& n_2, n_1) + (m_3, m_2, m_1) \\
        &= \big(n_3 + m_3 + f(n_2, m_2) + f(n_2 + m_2, f(n_1, m_1), n_2 + m_2 + f(n_1, m_1)), n_1 + m_1\big),
\end{align*}
and denote the resulting group by $(\Z_b)_f^3$ for 2-equivariant cocycle $f$.

Redefining $\phi_{s_2}: (\Z_b)_f^3 \to \Z_{b^3}$, $\hat{\imath}_2: \Z_b \times_f \Z_b \to (\Z_b)_f^3$, and $\hat{\pi}_2: (\Z_b)_f^3 \to \Z_b$ as 
\begin{gather*}
    \phi_{s_2}(n_3, n_2, n_1) = i_2(\phi_s(n_3, n_2)) + s_2(n_1) = i_2(i(n_3)) + i_2(s(n_2)) + s_2(n_1), \\
    \hat{\imath}_2(n_2, n_1) = (n_2, n_1, 0), \\
    \hat{\pi}_2(n_3, n_2, n_1) = n_1,
\end{gather*}
then $(\Z_b)_f^3$ is equivalent to $\Z_{b^3}$ defined entirely in terms of 2-equivariant cocycle $f$.

Finally, we define the $k$-digit base representation $(\Z_b)_f^k$ for any $k \in \N$, which allows us to represent arbitrary $n \in \Z$ as a sequence of digits $(n_k, n_{k-1}, ..., n_1)$.
First, let $(\Z_b)_f^k$ be the product $(\Z_b)^k$ with addition defined by
\begin{equation} \label{eq:group-law-recursive}
    n_j + m_j + f(n_{j-1}, m_{j-1}) + f(n_{j-1} + m_{j-1}, c_{j-1})
\end{equation}
for digits $j \in [k]$, where $c_j = f(n_{j-1}, m_{j-1}) + f(n_{j-1} + m_{j-1}, c_{j-1})$ for $j \in [k]$, and $c_0 = n_0 = m_0 = 0$, and $f: \Z_b \times \Z_b \to \Z_b$ is some function.
If $f$ is a \textit{$k$-equivariant cocycle}, i.e., a cocycle of $\Z_b$ with coefficients in $\Z_b$ such that $(\Z_b)_f^{k+1}$ forms a group with the addition (\ref{eq:group-law-recursive}) -- or, in group cohomological terms, such that $f_k = (0, ..., 0, f)$ is a cocycle of $\Z_b$ with coefficients in $(\Z_b)_f^k$ -- then $(\Z_b)_f^k$ forms a group.
Notice this reduces exactly to $(\Z_b)_f^2 = \Z_b \times_f \Z_b$ and $(\Z_b)_f^3$ for $k=2$ and 3, respectively.
It is easy to show by induction, with a similar diagram to Figure~\ref{fig:A10}, that if $(\Z_b)_f^k$ is equivalent to $\Z_{b^k}$ and $f$ is a $k$-equivariant cocycle, then $(\Z_b)_f^{k+1}$ is equivalent to $\Z_{b^{k+1}}$.
Simply define $\phi_{s_k}: (\Z_b)_f^{k+1} \to \Z_{b^{k+1}}$, $\hat{\imath}_k: (\Z_b)_f^k \to (\Z_b)_f^{k+1}$, and $\hat{\pi}_k: (\Z_b)_f^{k+1} \to \Z_b$ by
\begin{gather*}
    \phi_{s_k} = i_k\big(\phi_{s_{k-1}}(n_{k+1}, ..., n_2)\big) + s_k(n_1), \\
    \hat{\imath}_k(n_k, ..., n_1) = (n_k, ..., n_1, 0), \\
    \hat{\pi}_k(n_{k+1}, ..., n_1) = n_1,
\end{gather*}
where $s_k: \Z_b \to \Z_{b^{k+1}}$ is a section corresponding to cocycle $f_k = (0, ..., 0, f)$ and $i_k: \Z_{b^k} \to \Z_{b^{k+1}}$ is the usual inclusion map, and the diagram will commute as desired.
Therefore, we are able to faithfully represent any $n \in \Z$ by a sequence of digits $(n_k, n_{k-1}, ..., n_1)$, and so we have a complete base representation of the integers.

Note that, $(\Z_b)_f^k$ and $(\Z_b)_{f'}^k$ are equivalent if and only if $\Z_b \times_f \Z_b$ and $\Z_b \times_{f'} \Z_b$ are equivalent and $f$, $f'$ are $k$-equivariant.
So, for a given base representation $(\Z_b)_f^k$ and associated $f$, it suffices to check that $f'$ is $k$-equivariant and $f' = f + \delta c$ in order to show that $(\Z_b)_{f'}^k$ is equivalent to $(\Z_b)_f^k$.

\newpage

\section{Supplement to Section~\ref{sec:3}: Quantitative Measures of Carry Functions} \label{app:B}

\subsection{Additional Carry Tables} \label{app:B.1}

\begin{figure}[h]
    \centering
    \includegraphics[width=0.72\textwidth]{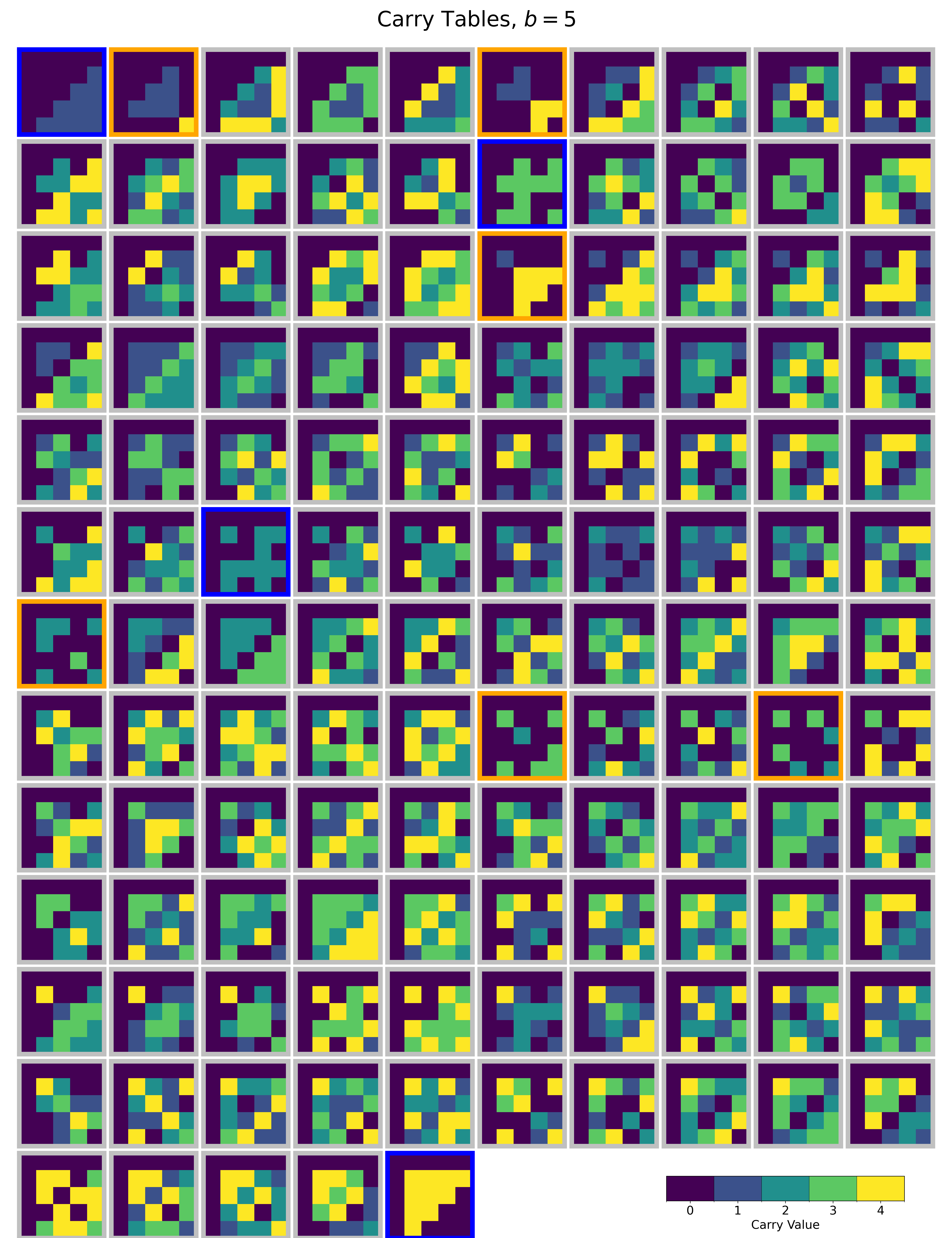}
    \caption{
    The 125 carry tables in base 5.
    The Single Value carry functions are outlined in blue (including the $\mathbf{1}$ carry function in the top left), the Low Dimensional Multiple Value carry functions in orange, and the Other Multiple Value carry functions in grey (see Section~\ref{sec:2.4}).
    Each table's entries are indexed by $\{0, 1, ...,e b-1\}$ from left to right, top to bottom; color indicates the value that is carried (see legend at bottom right).
    }
    \label{fig:B11}
\end{figure}

\newpage

\subsection{Comparison of Quantitative Measures} \label{app:B.2}

\begin{figure}[h]
    \centering
    \includegraphics[width=0.6\linewidth]{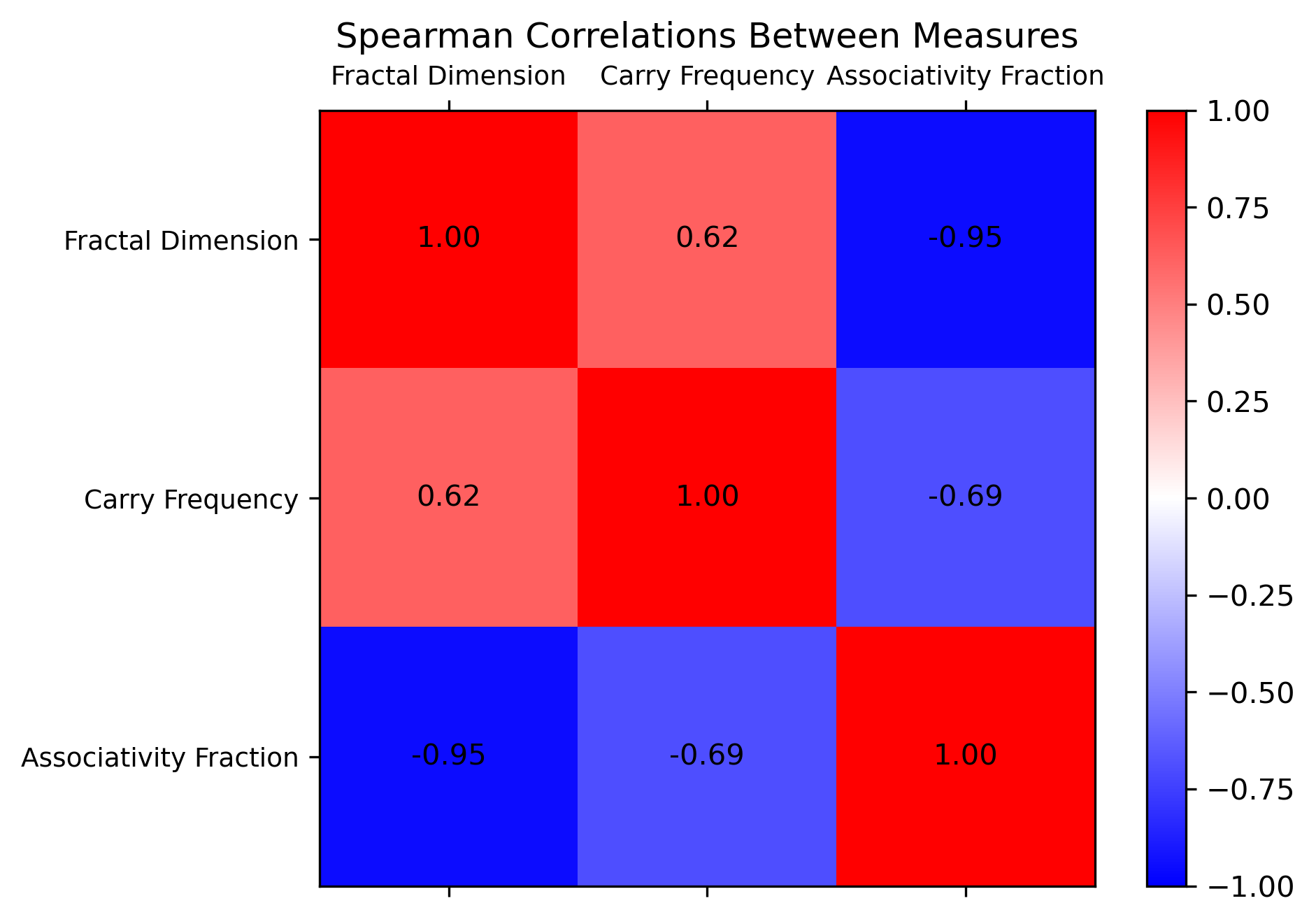}
    \caption{
    Spearman correlations between the three quantitative measures: fractal dimension, carry frequency, and associativity fraction.
    This demonstrates that the measures are not identical, and reflect different aspects of a carry function's structure.
    }
    \label{fig:B12}
\end{figure}

\newpage

\section{Supplement to Section~\ref{sec:4}: Neural Network Simulations} \label{app:C}

\subsection{LSTM Results} \label{app:C.1}

\begin{figure}[h]
    \centering
    \includegraphics[width=0.9\textwidth]{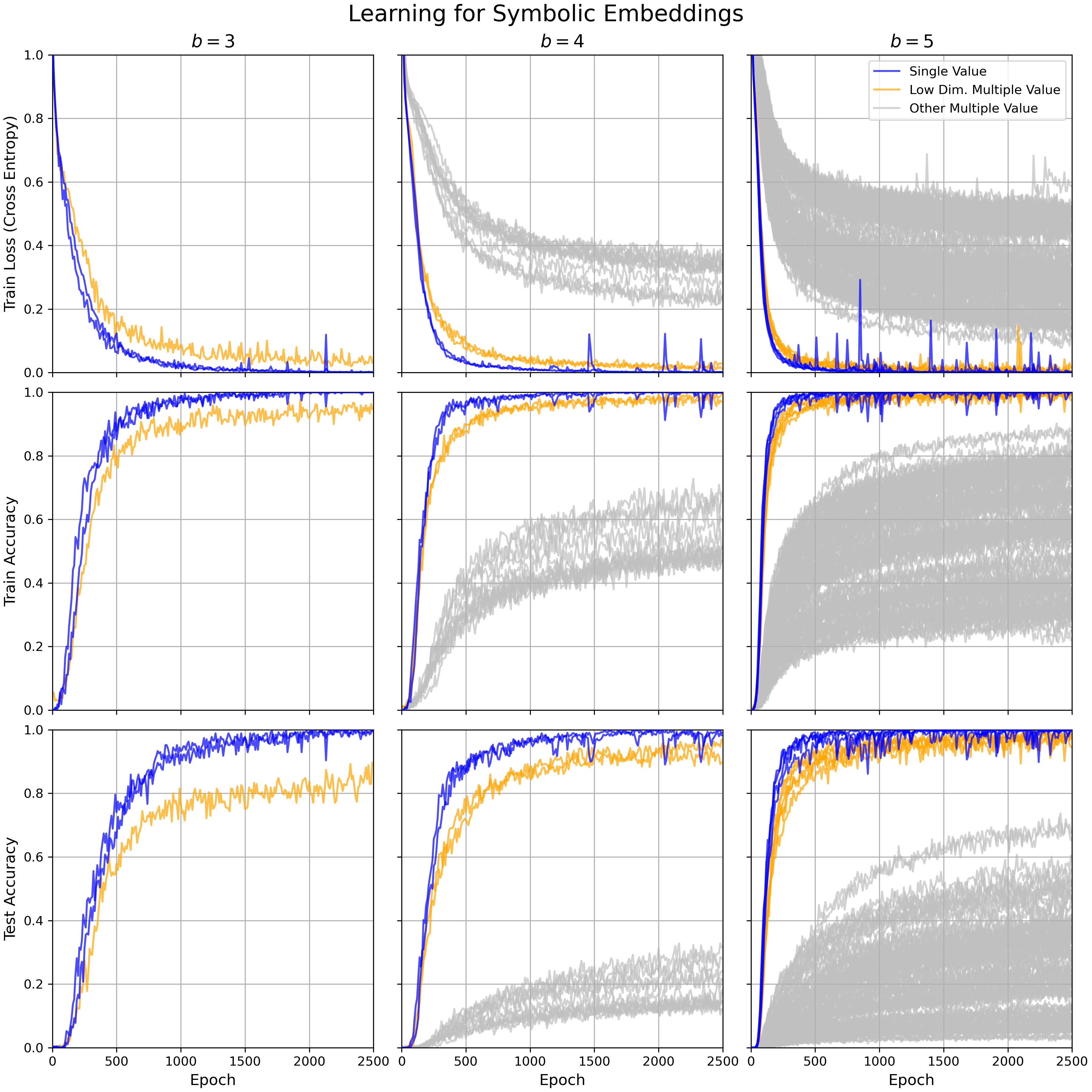}
    \caption{
    Learning for symbolic (one-hot) embeddings of digits using an LSTM.
    Performance over the course of training (averaged over 10 runs of each model implementation) for addition on different carry functions for bases $b=3$ to 5.
    }
    \label{fig:C13}
\end{figure}

\begin{figure}[p]
    \centering
    \includegraphics[width=0.6\textwidth]{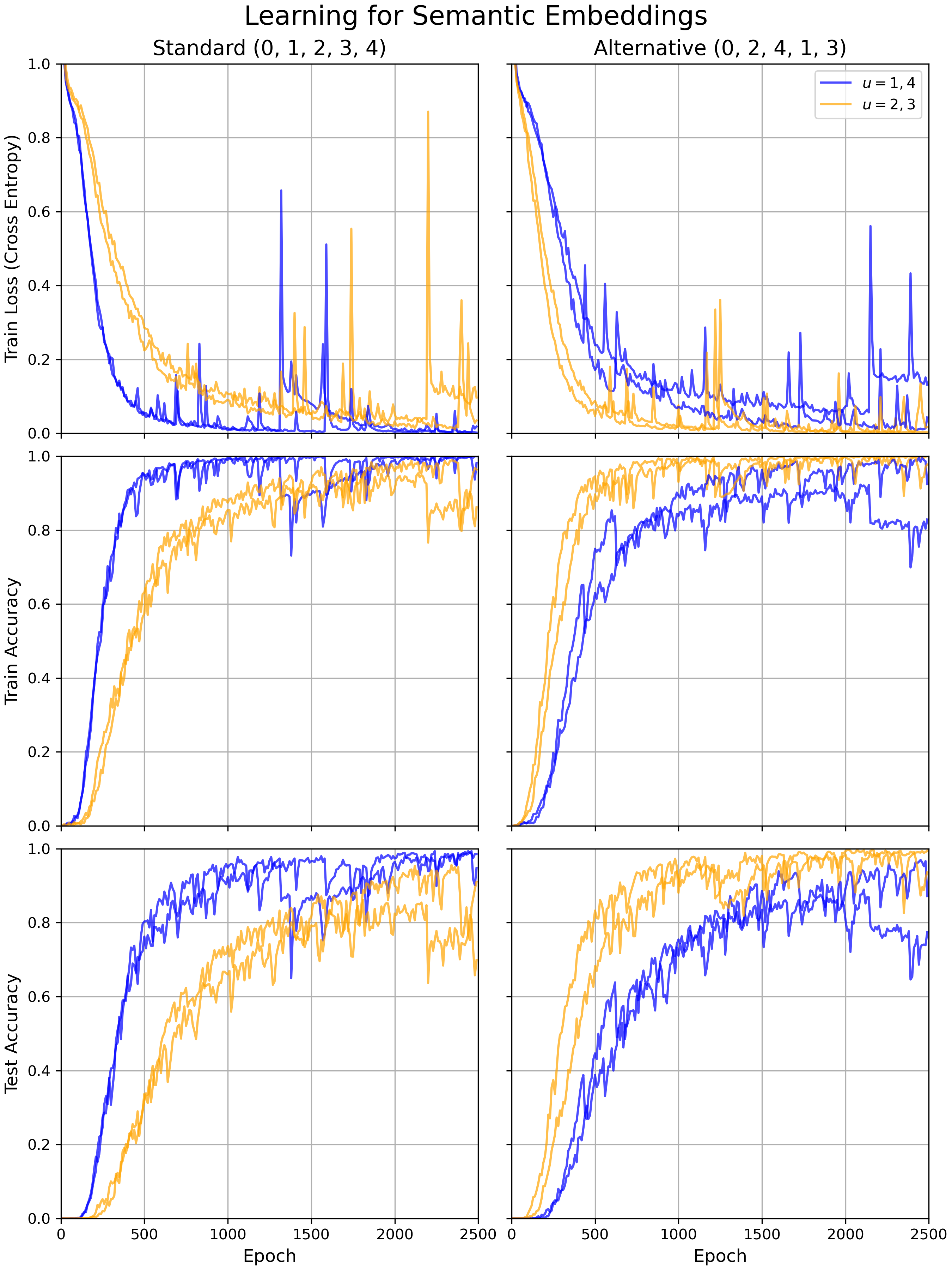}
    \caption{
    Learning for semantic embeddings of digits using an LSTM.
    Performance over the course of training on  Single Value carry functions with order-encoded inputs in the two non-degenerate orderings (0, 1, 2, 3, 4) and (0, 2, 4, 1, 3) for $b=5$ (see Section~\ref{sec:2.4.1} for how Single Value carry functions correspond to orderings of $\Z_b$).
    }
    \label{fig:C14}
\end{figure}

\begin{figure}[p]
    \centering
    \includegraphics[width=\textwidth]{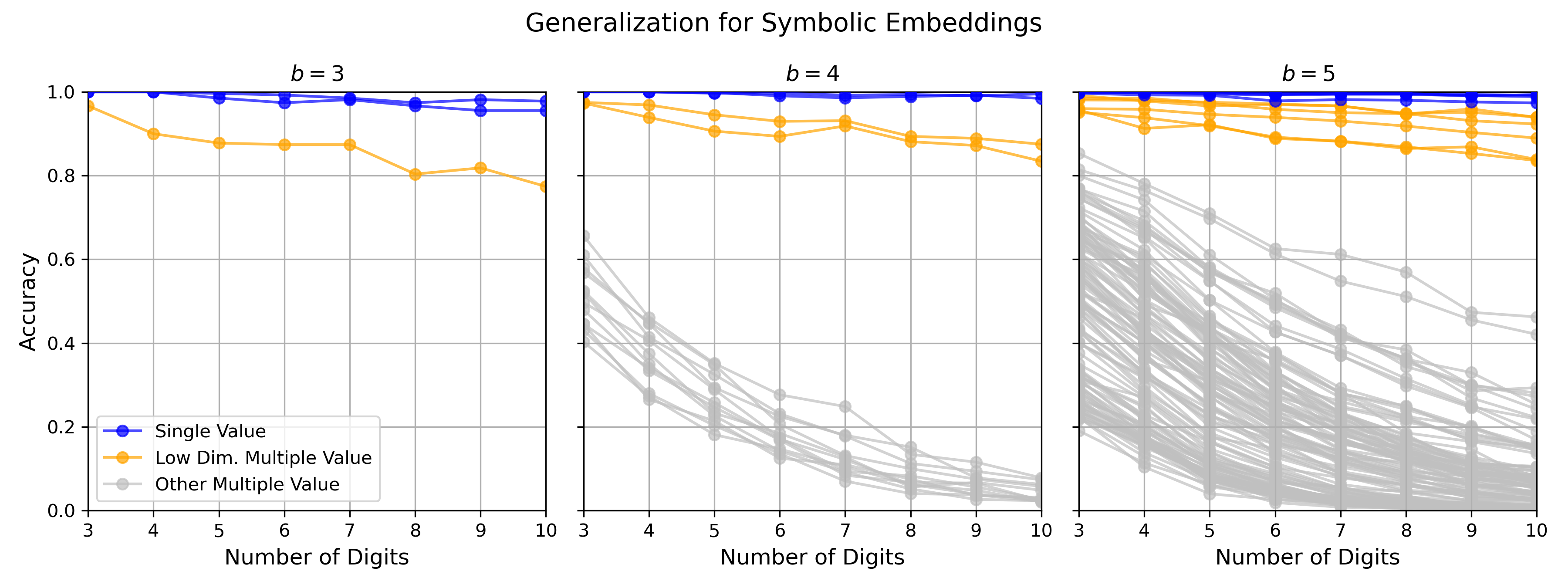}
    \caption{
    Out-of-domain generalization for symbolic embeddings of digits using an LSTM.
    For each carry function in bases $b=3$ to 5, accuracy was tested on $k$-digit numbers for each $k \in [3:10]$ after training on 3-digit numbers (averaged over 10 training/testing runs).
    }
    \label{fig:C15}

    \vspace{1cm}

    \includegraphics[width=0.9\textwidth]{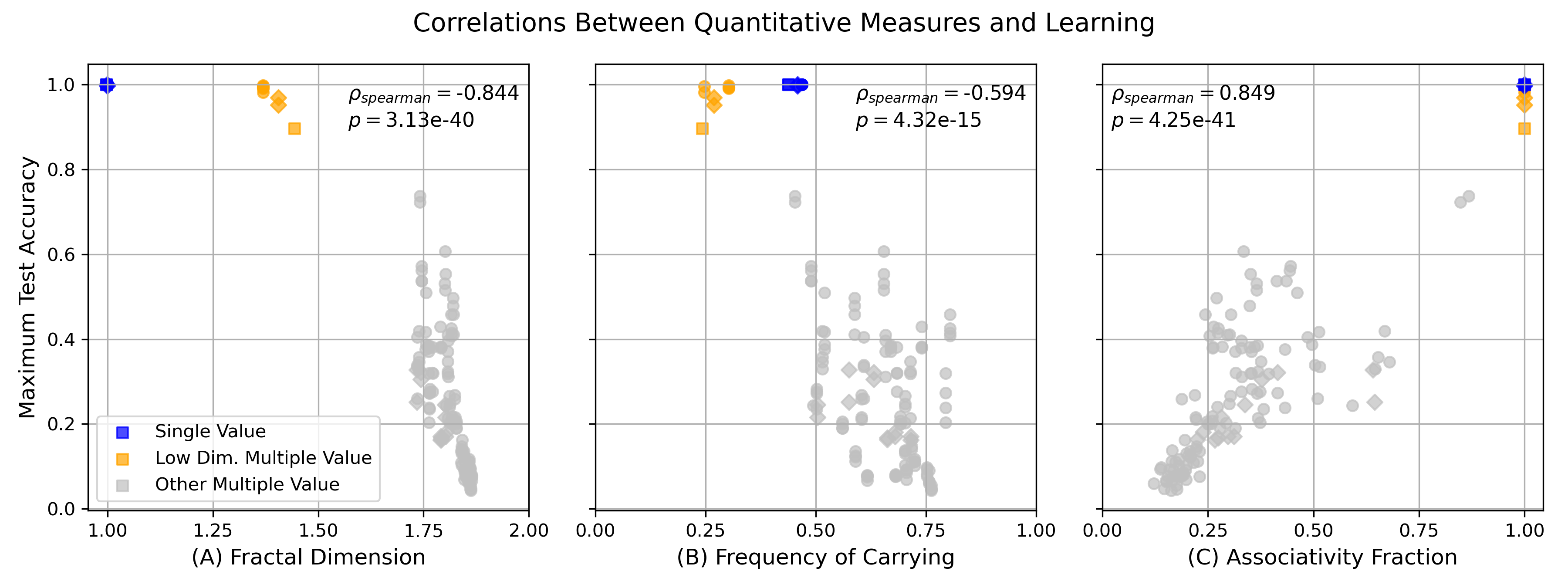}
    \caption{
    Relationship of carry function structure and learning using an LSTM, as measured by maximum testing accuracy (on 6-digit numbers).
    Scatter plots showing relationship of learning curves to structure measured as (A) fractal dimension, (B) frequency of carrying, and (C) associativity fraction for different carry functions, divided into three categories: Single Value carry functions (blue), Low  Dimensional Multiple Value carry functions (orange), and other Multiple Value carry functions (grey). 
    The base $b$ of each carry function is indicated by shape ($b=3$: squares; $b=4$: diamonds; and $b=5$: circles).
    Spearman's rank correlations (over bases $b=3$ to 5) and significance are shown for each plot.
    }
    \label{fig:C16}
\end{figure}

Here we present the same results as in Section~\ref{sec:4}, but using a single-layer LSTM in place of a GRU.
Otherwise, the training procedure was identical.
Figures~\ref{fig:C13} to \ref{fig:C16} show the training results for symbolic (one-hot) embeddings and semantic embeddings, the generalization results, and scatter plots of maximum testing accuracy (on 6-digit numbers) and the three quantitative measures.

\newpage

\subsection{Sigmoid Fit} \label{app:C.2}

\begin{figure}[h]
    \centering
    \includegraphics[width=0.5\textwidth]{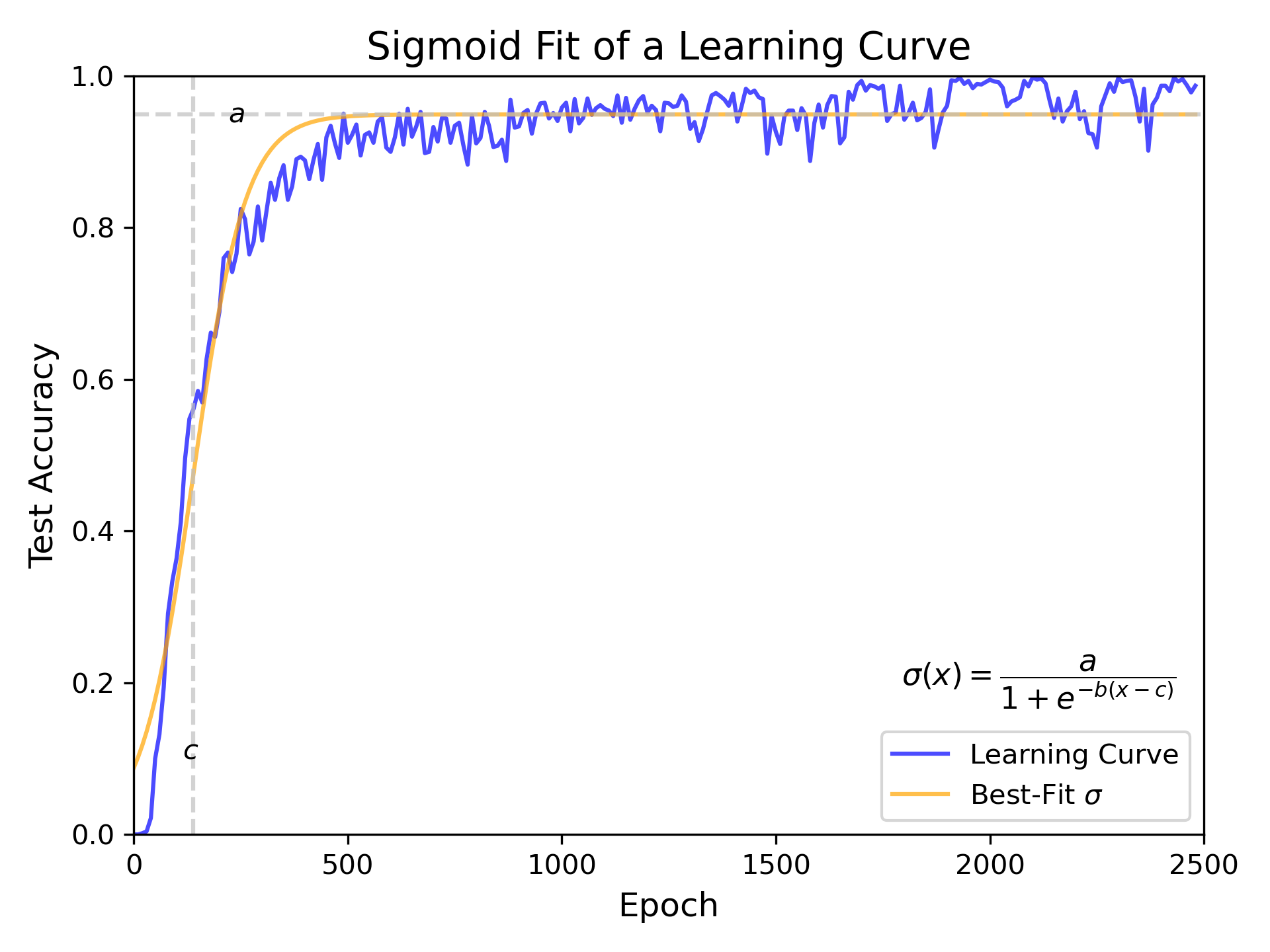}

    \includegraphics[width=0.5\textwidth]{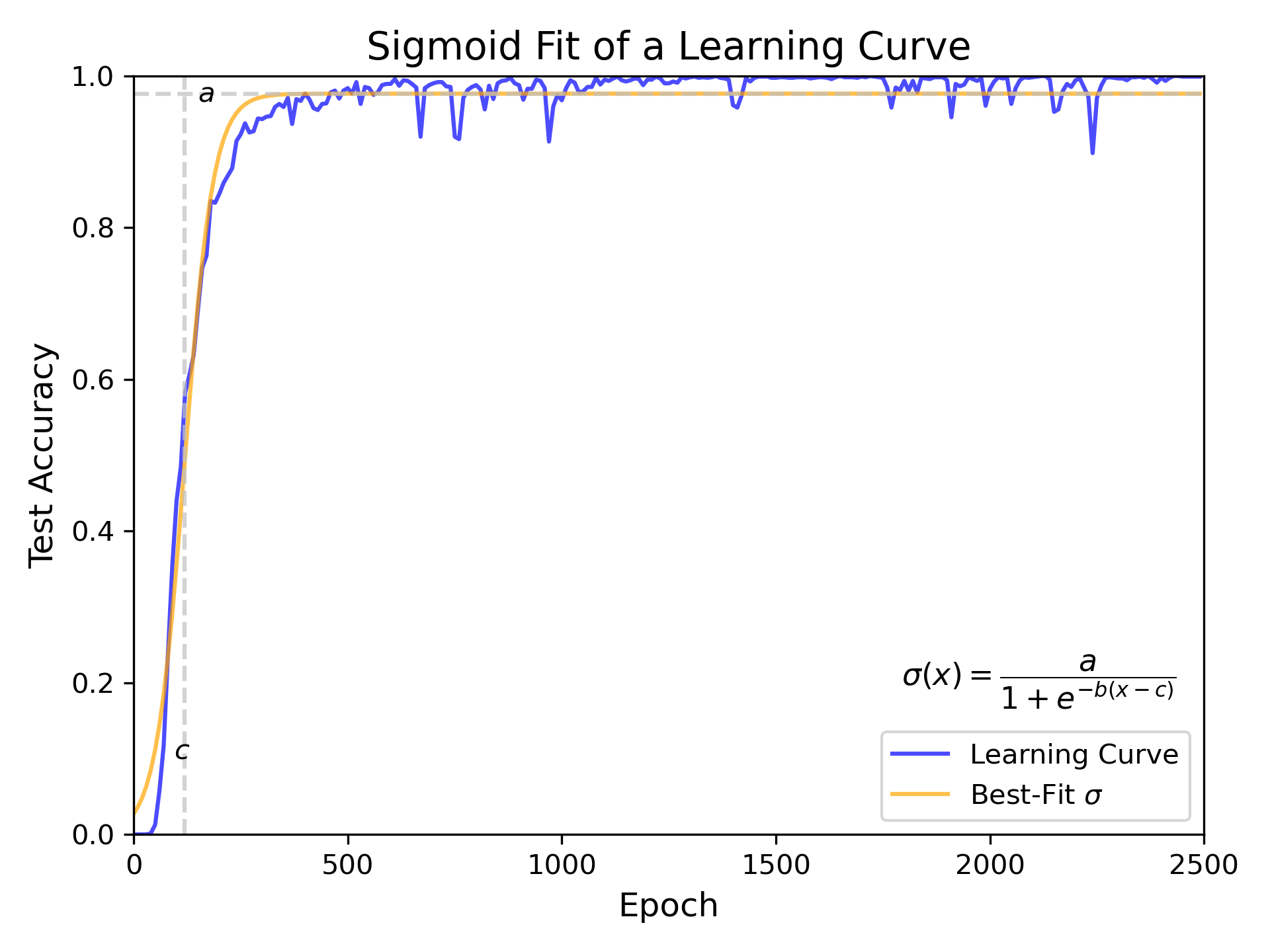}
    \caption{
    Sigmoid fit to learning curve of a GRU (top) and LSTM (bottom).
    A well-fit example of the sigmoid function $\sigma$ fit to a test accuracy curve (top: $R^2 = 0.96$; bottom: $R^2 = 0.98$), and the corresponding upper asymptote ($a$) and critical point ($c$).
    }
    \label{fig:C17}
\end{figure}

For robustness, we also quantified learning by fitting a sigmoid function $\sigma(x) = \frac{a}{1 + e^{-b(x-c)}}$, with parameters of upper asymptote $a$, growth rate $b$, and critical point $c$, to each test accuracy curve (on 6-digit numbers).
To avoid overfitting to fluctuations in the tails of the test accuracy curves, we restricted our fit to the portion of the curve up until it reached its maximum observed value.
Curve fitting was performed using non-linear least squares via \texttt{scipy.optimize.curve\_fit}.
For both GRUs and LSTMs, the sigmoid provided a good fit to test accuracy curves ($R^2 = 0.84 \pm 0.09$ and $R^2 = 0.92 \pm 0.04$, respectively).
Figure~\ref{fig:C17} shows well-fit test accuracy curves for a GRU (top) and LSTM (bottom).

\begin{figure}
    \centering
    \includegraphics[width=0.9\textwidth]{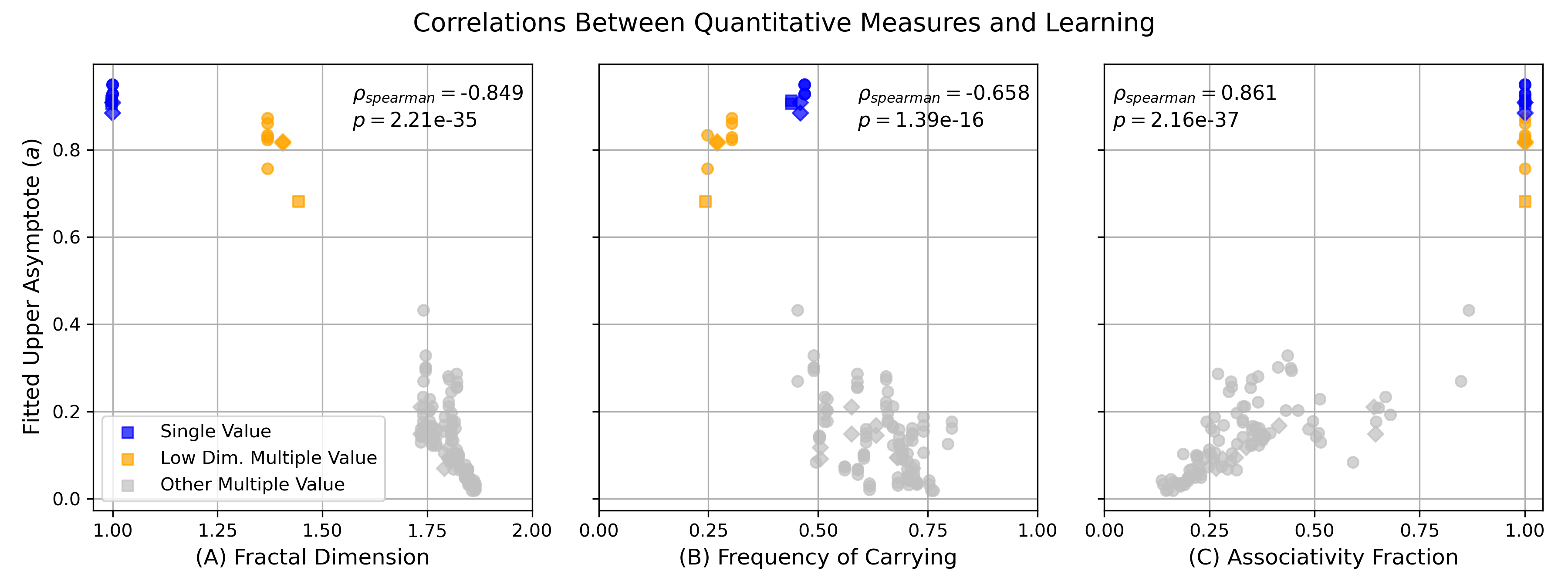}
    
    \includegraphics[width=0.9\textwidth]{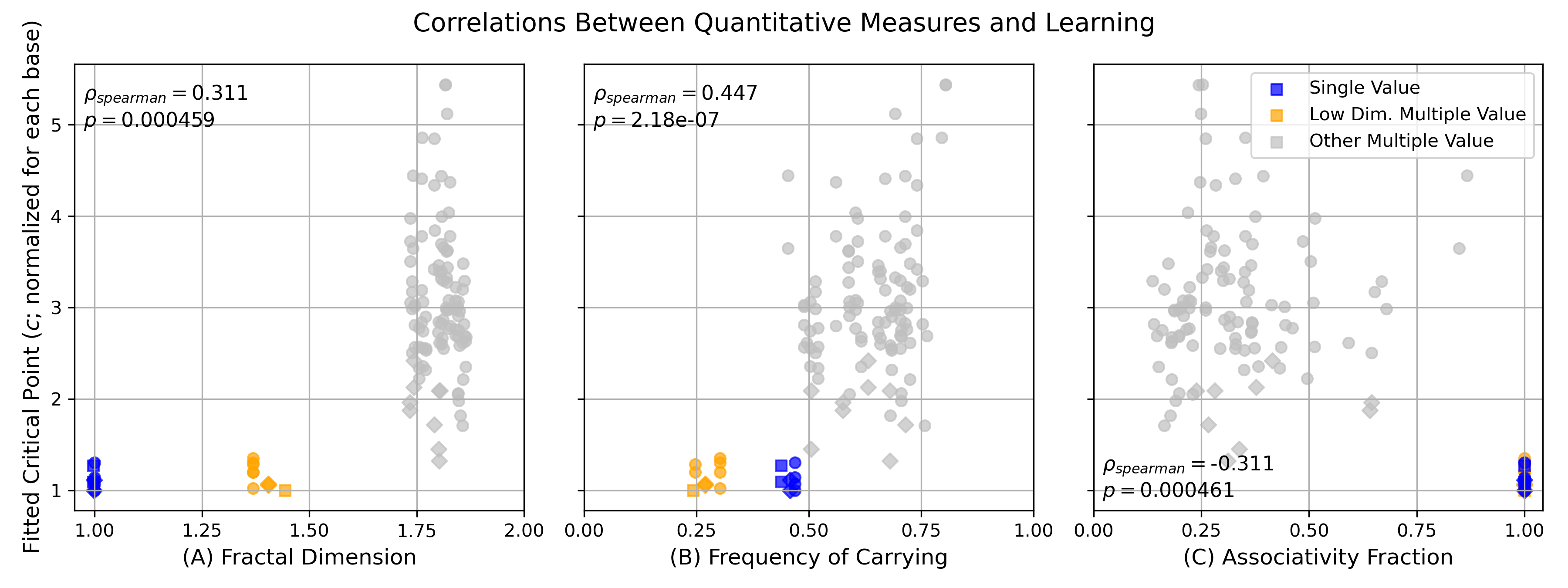}
    \caption{
    Relationship of carry function structure and learning using a GRU, as measured by the best-fit sigmoid's upper asymptote (top) and critical point (bottom).
    Scatter plots showing relationship of learning curves to structure measured as (A) fractal dimension, (B) frequency of carrying, and (C) associativity fraction for different carry functions, divided into three categories: Single Value carry functions (blue), Low  Dimensional Multiple Value carry functions (orange), and other Multiple Value carry functions (grey). 
    The base $b$ of each carry function is indicated by shape ($b=3$: squares; $b=4$: diamonds; and $b=5$: circles).
    Spearman's rank correlations (over bases $b=3$ to 5) and significance are shown for each plot.
    Note that \texttt{scipy.optimize.curve\_fit} was unable to learn some learning curves, which are excluded from the figure ($\sim 14.6\%$).
    }
    \label{fig:C18}
\end{figure}

\begin{figure}
    \centering
    \includegraphics[width=0.9\textwidth]{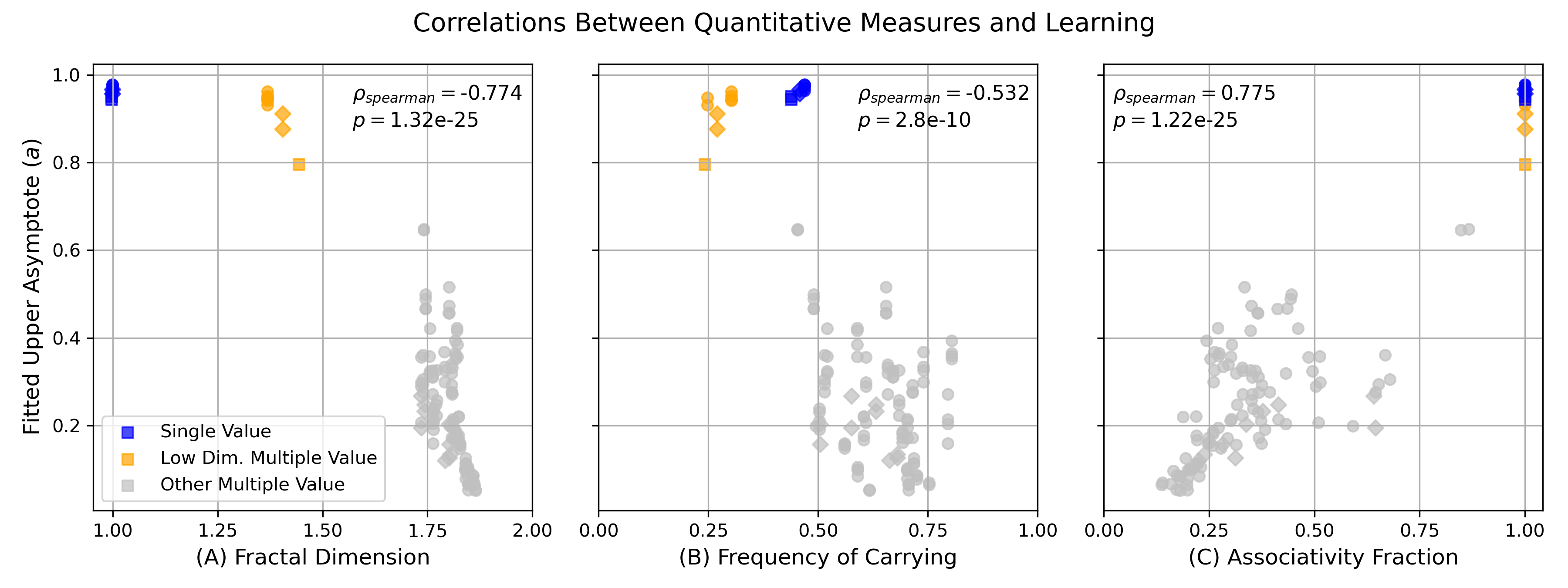}
    
    \includegraphics[width=0.9\textwidth]{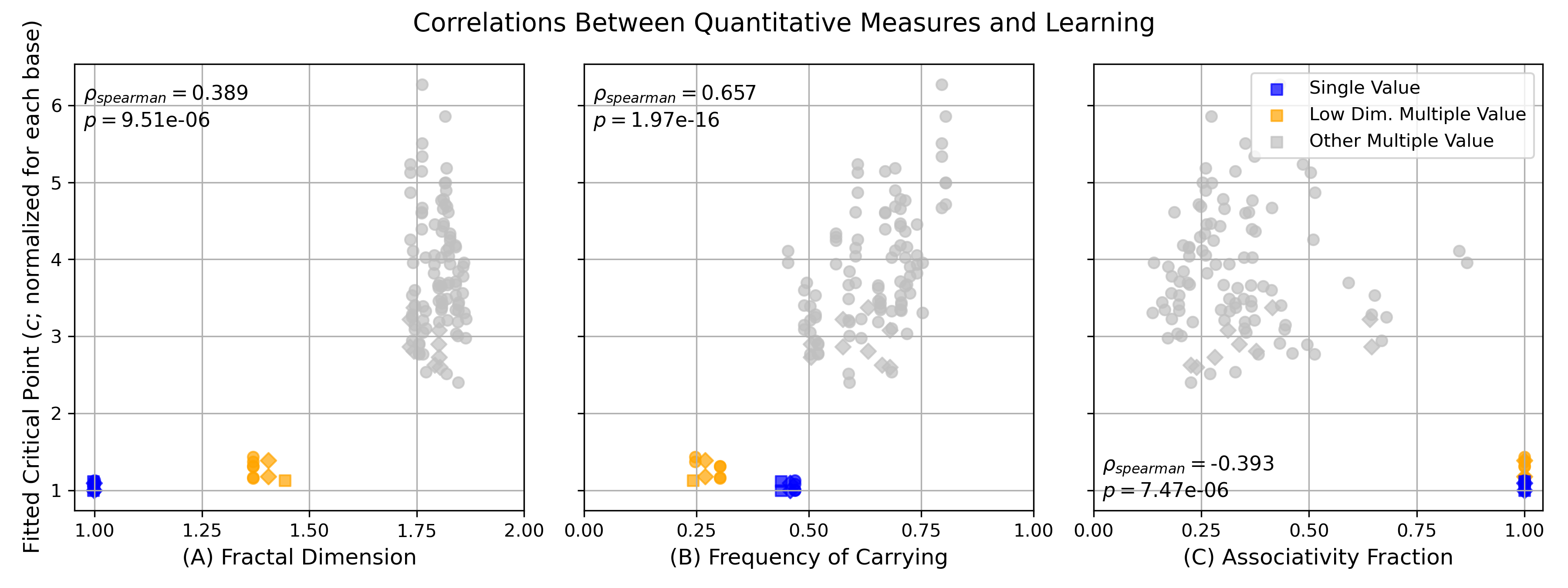}
    \caption{
    Relationship of carry function structure and learning using an LSTM, as measured by the best-fit sigmoid's upper asymptote (top) and critical point (bottom).
    Scatter plots showing relationship of learning curves to structure measured as (A) fractal dimension, (B) frequency of carrying, and (C) associativity fraction for different carry functions, divided into three categories: Single Value carry functions (blue), Low  Dimensional Multiple Value carry functions (orange), and other Multiple Value carry functions (grey). 
    The base $b$ of each carry function is indicated by shape ($b=3$: squares; $b=4$: diamonds; and $b=5$: circles).
    Spearman's rank correlations (over bases $b=3$ to 5) and significance are shown for each plot.
    Note that \texttt{scipy.optimize.curve\_fit} was unable to learn some learning curves, which are excluded from the figure ($\sim 15.3\%$).
    }
    \label{fig:C19}
\end{figure}

From the best-fit sigmoid, we took two measurements of learning: its upper asymptote and critical point.
For the critical points specifically, we divisively normalized by the minimal critical point for each base (corresponding to the fastest-learned carry function), in order to examine the structure of carry functions independent of base.
Figures~\ref{fig:C18} and \ref{fig:C19} show scatter plots of these measurements of learning speed and the three quantitative measures using a GRU and LSTM, respectively.

\end{document}